\documentclass{article}

\usepackage{microtype}
\usepackage{graphicx}
\usepackage{subfigure}
\usepackage{booktabs} 

\usepackage{hyperref}
\usepackage{url}
\usepackage{subfiles}
\usepackage{listing}
\usepackage{dsfont}
\usepackage{color}
\usepackage{wrapfig,lipsum,booktabs}
\usepackage{tikz}
\usepackage{xfrac}
\usepackage{floatrow}
\usepackage{xcolor,colortbl}
\usepackage{multirow}
\usepackage{graphicx}
\usepackage{enumitem}
\usepackage{amsmath}
\usepackage{arydshln}
\usepackage{float}
\usepackage{amssymb}
\usepackage{caption}
\usepackage{rotating}

\usepackage{hyperref}

\newcommand{\MaxPool}{\text{MaxPool}}
\newcommand{\AvgPool}{\text{AvgPool}}
\newcommand{\Downsample}{\text{Subsample}}

\newcommand{\Conv}{\text{Conv}}
\newcommand{\Relu}{\text{Relu}}
\newcommand{\Blur}{\text{Blur}}
\newcommand{\Max}{\text{Max}}
\newcommand{\BlurDown}{\text{BlurPool}}
\newcommand{\Shift}{\text{Shift}}

\definecolor{red}{RGB}{255,0,0}
\definecolor{green}{RGB}{10,150,30}
\definecolor{blue}{RGB}{0,0,255}
\definecolor{magenta}{RGB}{160,0,200}
\definecolor{gray}{RGB}{96,96,96}
\definecolor{darkred}{RGB}{180,50,20}
\definecolor{darkgreen}{RGB}{20,180,50}
\definecolor{purple}{RGB}{90,20,90}
\definecolor{darkblue}{RGB}{59,76,192}
\definecolor{lightyellow}{RGB}{255,253,141}
\definecolor{lightgreen}{RGB}{200, 249, 182}
\definecolor{lightred}{RGB}{242,139,128}

\newcommand{\best}[1]{{#1}}
\newcommand{\worst}[1]{{#1}}

\usepackage[accepted]{icml2019}

\icmltitlerunning{Making Convolutional Networks Shift-Invariant Again}

\begin{document}

\twocolumn[
\icmltitle{Making Convolutional Networks Shift-Invariant Again}




\begin{icmlauthorlist}
\icmlauthor{Richard Zhang}{adobe}
\end{icmlauthorlist}

\icmlaffiliation{adobe}{Adobe Research, San Francisco, CA}

\icmlcorrespondingauthor{Richard Zhang}{rizhang@adobe.com}

\icmlkeywords{Convolutional networks, translation-invariance, shift-invariance, anti-aliasing, anti-aliased}

\vskip 0.3in
]



\printAffiliationsAndNotice{}  

\begin{abstract}

Modern convolutional networks are not shift-invariant, as small input shifts or translations can cause drastic changes in the output. Commonly used downsampling methods, such as max-pooling, strided-convolution, and average-pooling, ignore the sampling theorem. The well-known signal processing fix is anti-aliasing by low-pass filtering before downsampling.
However, simply inserting this module into deep networks degrades performance; as a result, it is seldomly used today. We show that when integrated correctly, it is compatible with existing architectural components, such as max-pooling and strided-convolution.
We observe \textit{increased accuracy} in ImageNet classification, across several commonly-used architectures, such as ResNet, DenseNet, and MobileNet, indicating effective regularization. Furthermore, we observe \textit{better generalization}, in terms of stability and robustness to input corruptions. Our results demonstrate that this classical signal processing technique has been undeservingly overlooked in modern deep networks.
	
\end{abstract}

\vspace{-7mm}
\section{Introduction}

When downsampling a signal, such an image, the textbook solution is to anti-alias by low-pass filtering the signal~\cite{oppenheim99dtsp, gonzalez92imageproc}. Without it, high-frequency components of the signal alias into lower-frequencies. This phenomenon is commonly illustrated in movies, where wheels appear to spin backwards, known as the Stroboscopic effect, due to the frame rate not meeting the classical sampling criterion~\cite{nyquist1928certain}. Interestingly, most modern convolutional networks do not worry about anti-aliasing.

\begin{figure*}[t]
\centering
\hspace{4mm}
\includegraphics[width=.48\linewidth,trim=0 65mm 0 0,clip]{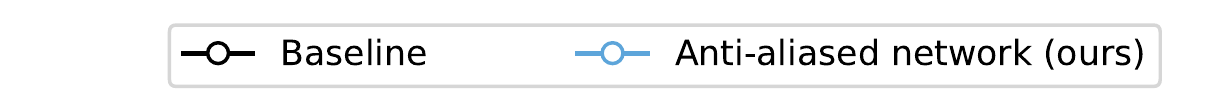}
\hspace{-2mm}
\includegraphics[width=.48\linewidth,trim=0 65mm 0 0,clip]{figures/heartbeat_template.pdf}
\\
\begin{turn}{90}
\hspace{1mm} {\bf AlexNet on ImageNet}
\end{turn}
\includegraphics[width=.48\linewidth,trim=0 11mm 0 10mm,clip]{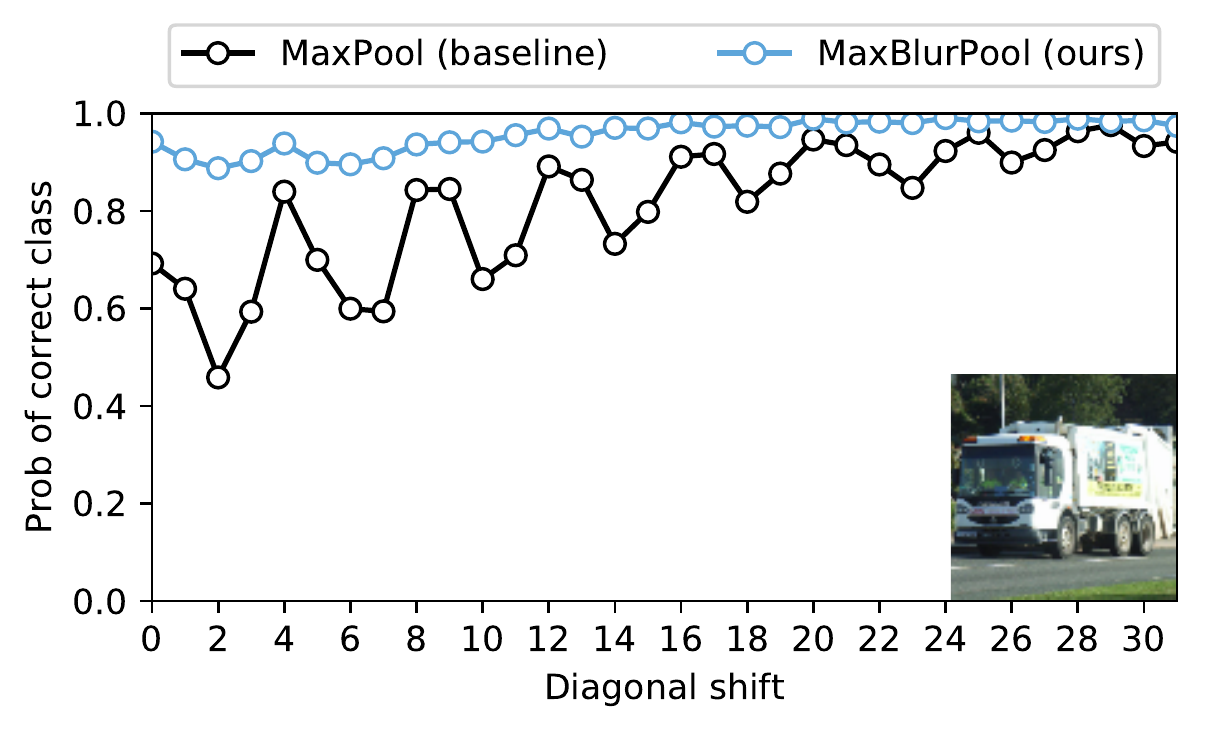}
\hspace{-2mm}
\includegraphics[width=.48\linewidth,trim=0 11mm 0 10mm,clip]{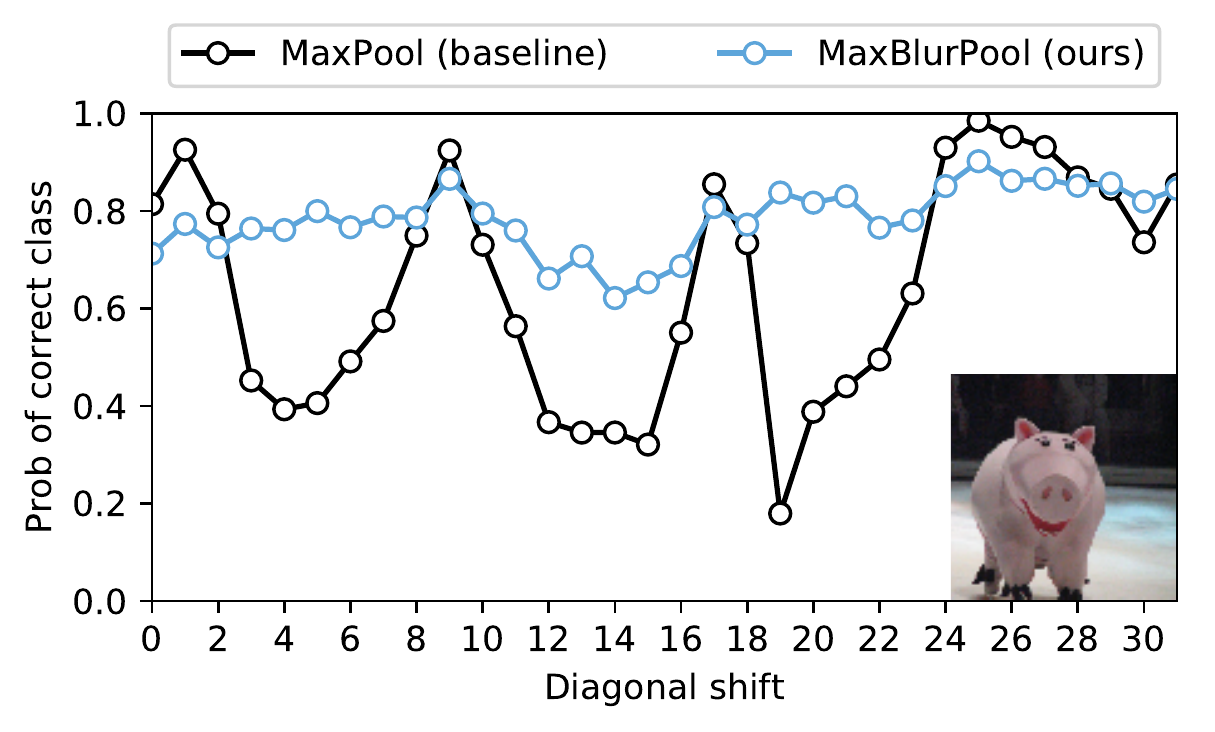}
\\

\begin{turn}{90}
\hspace{12mm} {\bf VGG on CIFAR}
\end{turn}
\includegraphics[width=.48\linewidth,trim=0 0 0 1cm,clip]{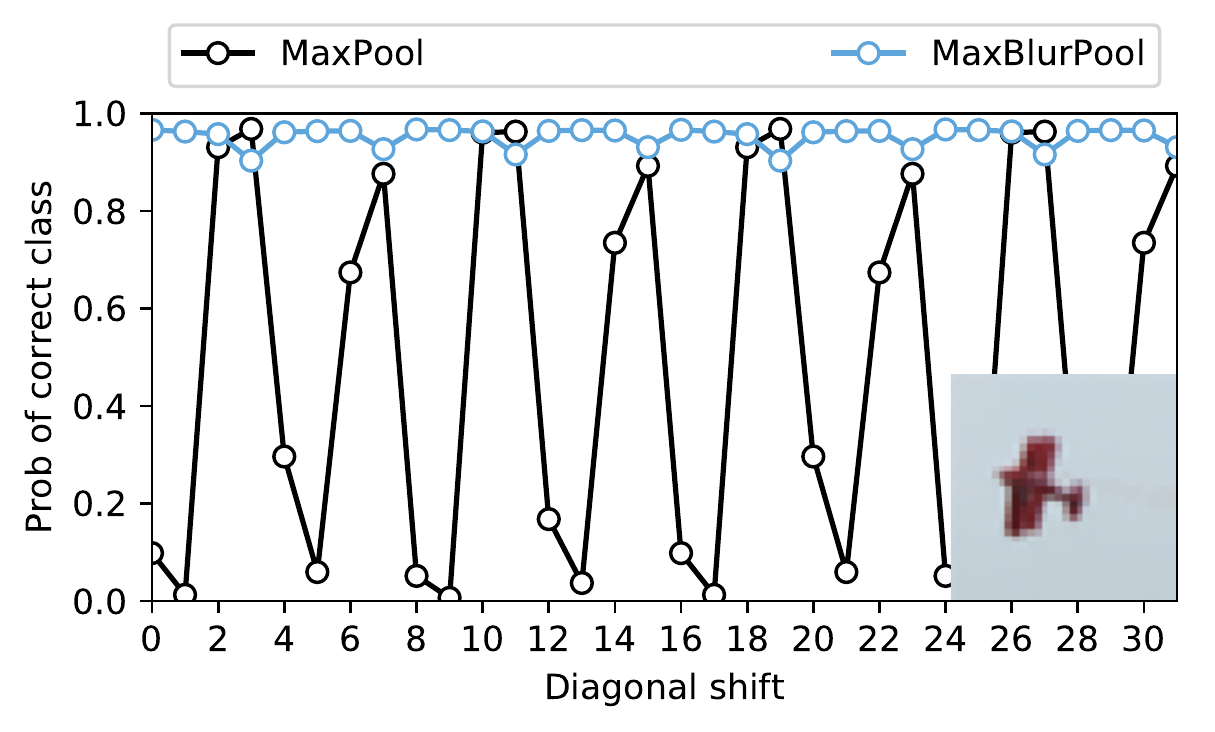}
\hspace{-2mm}
\includegraphics[width=.48\linewidth,trim=0 0 0 1cm,clip]{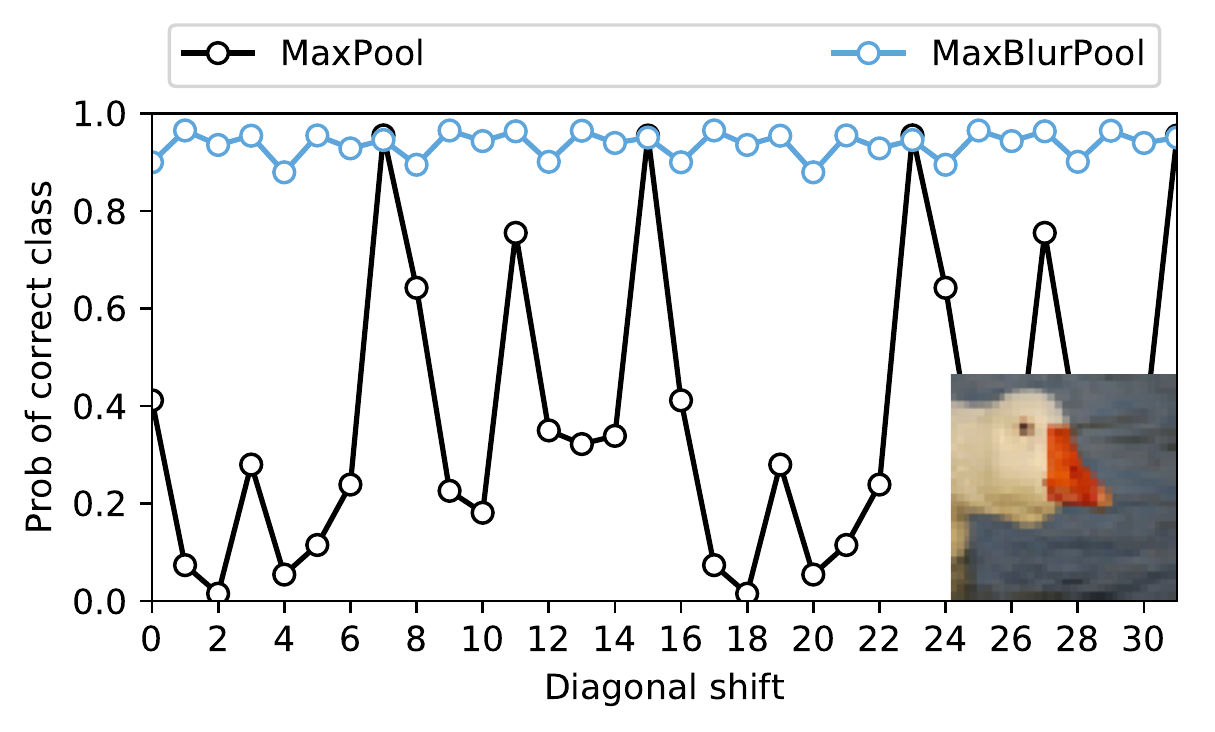} \\
\vspace{-6mm}
\caption{
{\bf Classification stability for selected images.} Predicted probability of the correct class changes when shifting the image. The baseline (black) exhibits chaotic behavior, which is stabilized by our method (blue). We find this behavior across networks and datasets. Here, we show selected examples using AlexNet on ImageNet {\bf (top)} and VGG on CIFAR10 {\bf (bottom)}. Code and anti-aliased versions of popular networks are available at \url{https://richzhang.github.io/antialiased-cnns/}.
\label{fig:class_stab}
\vspace{-3mm}
}
\end{figure*}

Early networks did employ a form of blurred-downsampling -- average pooling~\citep{lecun1990handwritten}. However, ample empirical evidence suggests max-pooling provides stronger task performance~\citep{scherer2010evaluation}, leading to its widespread adoption. Unfortunately, max-pooling does not provide the same anti-aliasing capability, and a curious, recently uncovered phenomenon emerges -- small shifts in the input can drastically change the output~\cite{engstrom2017rotation, azulay2019deep}. As seen in Figure~\ref{fig:class_stab}, network outputs can oscillate depending on the input position. 

Blurred-downsampling and max-pooling are commonly viewed as competing downsampling strategies~\cite{scherer2010evaluation}. However, we show that they are compatible. Our simple observation is that max-pooling is inherently composed of two operations: (1) evaluating the max operator densely and (2) naive subsampling. We propose to low-pass filter between them as a means of anti-aliasing. This viewpoint enables low-pass filtering to augment, rather than replace max-pooling. As a result, shifts in the input leave the output relatively unaffected (shift-invariance) and more closely shift the internal feature maps (shift-equivariance).

Furthermore, this enables proper placement of the low-pass filter, directly before subsampling. With this methodology, practical anti-aliasing can be achieved with any existing strided layer, such as strided-convolution, which is used in more modern networks such as ResNet~\cite{He_2016_CVPR} and MobileNet~\cite{sandler2018mobilenetv2}.

A potential concern is that overaggressive filtering can result in heavy loss of information, degrading performance. However, we actually observe \textit{increased accuracy} in ImageNet classification~\cite{russakovsky2015imagenet} across architectures, as well as \textit{increased robustness} and \textit{stability} to corruptions and perturbations~\cite{hendrycks2019using}.
In summary:

\vspace{-1mm}
\begin{itemize}[noitemsep,nolistsep,leftmargin=.2cm]
	\item We integrate classic anti-aliasing to improve shift-equivariance of deep networks. Critically, the method is compatible with existing downsampling strategies.
	\item We validate on common downsampling strategies -- max-pooling, average-pooling, strided-convolution -- in different architectures. We test across multiple tasks -- image classification and image-to-image translation.
	\item For ImageNet classification, we find, surprisingly, that accuracy increases, indicating effective regularization.
	\item Furthermore, we observe better generalization. Performance is more robust and stable to corruptions such as rotation, scaling, blurring, and noise variants.
\end{itemize}

\vspace{-4mm}
\section{Related Work}

Local connectivity and weight sharing have been a central tenet of neural networks, including the Neocognitron~\citep{fukushima1982neocognitron}, LeNet~\citep{lecun1998gradient} and modern networks such as Alexnet~\citep{krizhevsky2012imagenet}, VGG~\citep{simonyan2014very}, ResNet~\citep{He_2016_CVPR}, and DenseNet~\citep{huang2017densely}. In biological systems, local connectivity was famously discovered in a cat's visual system~\cite{hubel1962receptive}. Recent work has strived to add additional invariances, such as rotation, reflection, and scaling~\citep{sifre2013rotation, bruna2013invariant, kanazawa2014locally, cohen2016group, worrall2017harmonic, esteves2017polar}. We focus on shift-invariance, which is often taken for granted.

\begin{figure*}[t]
\begin{center}
\includegraphics[width=1.\linewidth]{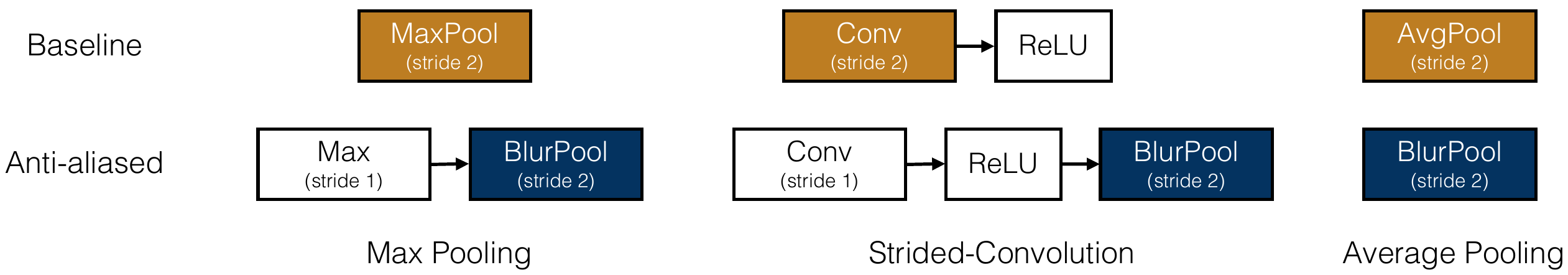}
\end{center}
\vspace{-5mm}
\caption{{\bf Anti-aliasing common downsampling layers.} {\bf (Top)} Max-pooling, strided-convolution, and average-pooling can each be better antialiased {\bf (bottom)} with our proposed architectural modification. An example on max-pooling is shown below.}
\label{fig:antialias-mods}
\vspace{-2mm}
\end{figure*}

\begin{figure*}[h]
\begin{center}
\includegraphics[width=1.\linewidth]{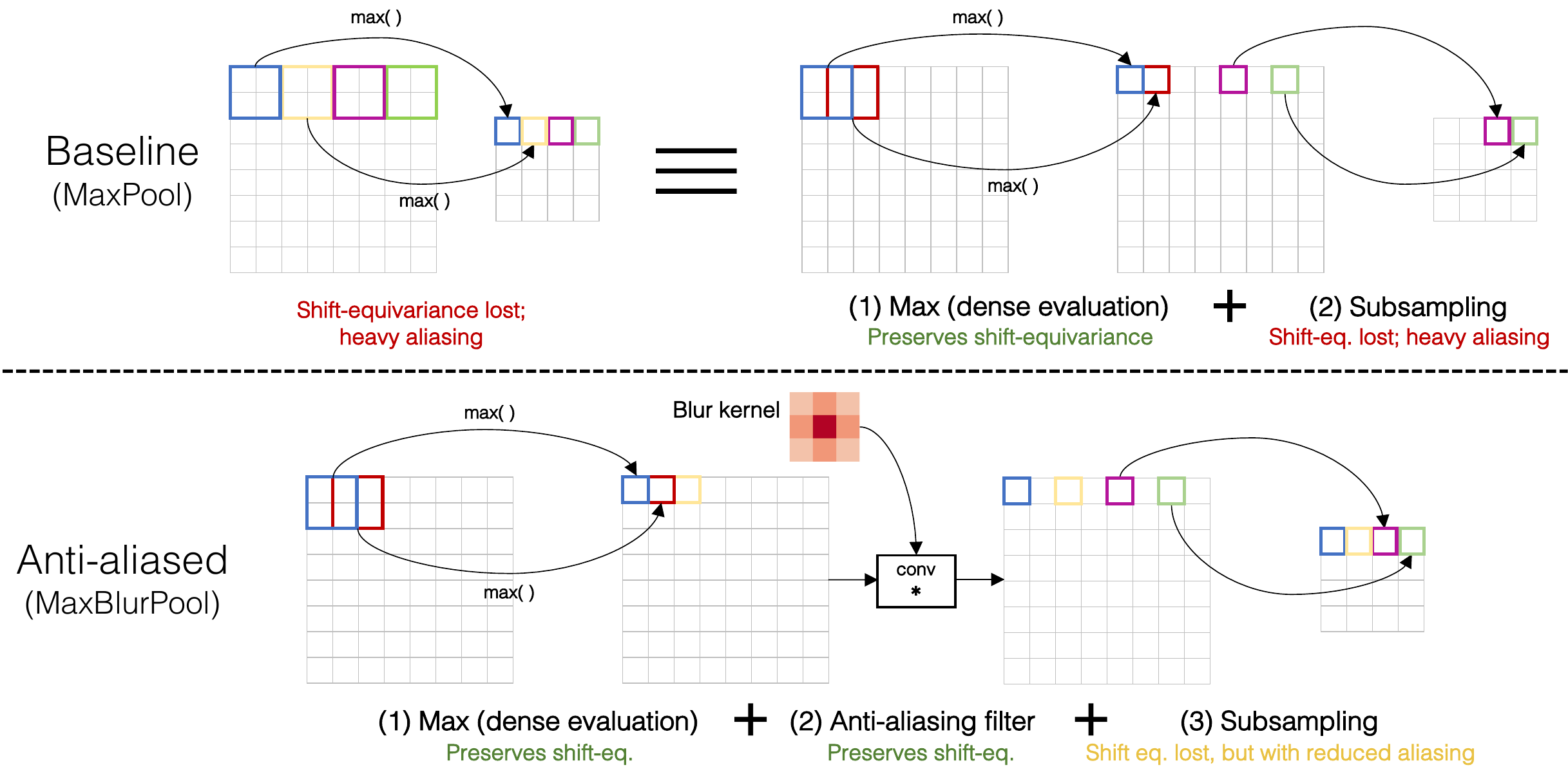}
\end{center}
\vspace{-5mm}
\caption{{\bf Anti-aliased max-pooling.} {\bf(Top)} Pooling does not preserve shift-equivariance. It is functionally equivalent to densely-evaluated pooling, followed by subsampling. The latter ignores the Nyquist sampling theorem and loses shift-equivariance. {\bf (Bottom)} We low-pass filter between the operations. This keeps the first operation, while anti-aliasing the appropriate signal. Anti-aliasing and subsampling can be combined into one operation, which we refer to as {\bf BlurPool}. 
\label{fig:fig1}
\vspace{-2mm}
}
\end{figure*}

Though different properties have been engineered into networks, what factors and invariances does an emergent representation actually learn? Qualitative analysis of deep networks have included showing patches which activate hidden units~\citep{girshick2014rich, zhou2014object}, actively maximizing hidden units~\citep{mordvintsev2015deepdream}, and mapping features back into pixel space~\citep{zeiler2014visualizing, henaff2015geodesics, mahendran2015understanding, dosovitskiy2016generating, dosovitskiy2016inverting, nguyen2017plug}. Our analysis is focused on a specific, low-level property and is complementary to these approaches.

A more quantitative approach for analyzing networks is measuring representation or output changes (or robustness to changes) in response to manually generated perturbations to the input, such as image transformations~\citep{goodfellow2009measuring, lenc2015understanding, azulay2019deep}, geometric transforms~\citep{fawzi2015manitest, ruderman2018pooling}, and CG renderings with various shape, poses, and colors~\citep{aubry2015understanding}. A related line of work is adversarial examples, where input perturbations are purposely directed to produce large changes in the output. These perturbations can be on pixels~\citep{goodfellow2014generative,goodfellow2014explaining}, a single pixel~\citep{su2017one}, small deformations~\citep{xiao2018spatially}, or even affine transformations~\citep{engstrom2017rotation}. We aim to make the network robust to the simplest of these types of attacks and perturbations: shifts. In doing so, we also observe increased robustness across other types of corruptions and perturbations~\cite{hendrycks2019using}.

\begin{figure*}[h]
\begin{center}
{\hspace{10mm} Baseline (MaxPool) \hspace{55mm} Anti-aliased (MaxBlurPool)}
\includegraphics[width=1.\linewidth,trim=0 .2cm 0 .2cm, clip]{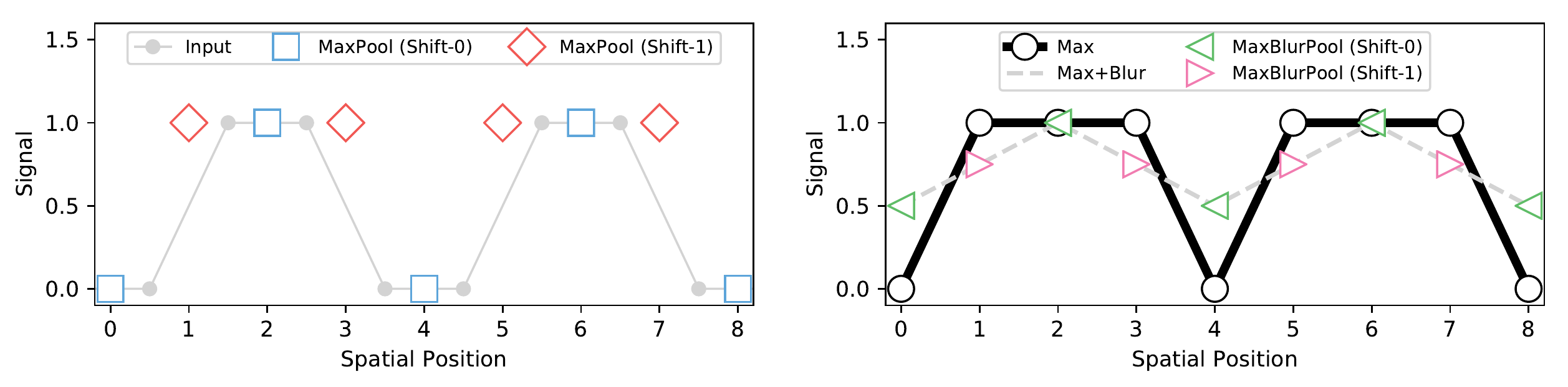}
\end{center}
\vspace{-5mm}
\caption{{\bf Illustrative 1-D example of sensitivity to shifts.} We illustrate how downsampling affects shift-equivariance with a toy example. {\bf (Left)} An input signal is in \textcolor{gray}{light gray line}. Max-pooled ($k=2$, $s=2$) signal is in \textcolor{blue}{blue squares}. Simply shifting the input and then max-pooling provides a completely different answer (\textcolor{red}{red diamonds}). {\bf (Right)} The blue and red points are subsampled from a densely max-pooled ($k=2$, $s=1$) intermediate signal ({\bf thick black line}). We low-pass filter this intermediate signal and then subsample from it, shown with \textcolor{green}{green} and \textcolor{magenta}{magenta} triangles, better preserving shift-equivariance.
\label{fig:simp_ex}
\vspace{-3mm}
}
\end{figure*}

Classic hand-engineered computer vision and image processing representations, such as SIFT~\citep{lowe1999object}, wavelets, and image pyramids~\citep{adelson1984pyramid,burt1987laplacian} also extract features in a sliding window manner, often with some subsampling factor. As discussed in~\citet{simoncelli1992shiftable}, literal shift-equivariance cannot hold when subsampling. Shift-equivariance can be recovered if features are extracted densely, for example textons~\citep{leung2001representing}, the Stationary Wavelet Transform~\citep{fowler2005redundant}, and DenseSIFT~\citep{vedaldi2010vlfeat}. Deep networks can also be evaluated densely, by removing striding and making appropriate changes to subsequent layers by using \emph{\'{a} trous}/dilated convolutions~\citep{chen2014semantic,chen2018deeplab,yu2015multi,yu2017dilated}. This comes at great computation and memory cost. Our work investigates improving shift-equivariance with minimal additional computation, by blurring before subsampling.

Early networks employed average pooling~\citep{lecun1990handwritten}, which is equivalent to blurred-downsampling with a box filter. However, work~\citep{scherer2010evaluation} has found max-pooling to be more effective, which has consequently become the predominant method for downsampling. While previous work~\cite{scherer2010evaluation,henaff2015geodesics,azulay2019deep} acknowledges the drawbacks of max-pooling and benefits of blurred-downsampling, they are viewed as separate, discrete choices, preventing their combination. Interestingly,~\citet{lee2016generalizing} does not explore low-pass filters, but does propose to softly gate between max and average pooling. However, this does not fully utilize the anti-aliasing capability of average pooling.

\citet{mairal2014convolutional} derive a network architecture, motivated by translation invariance, named Convolutional Kernel Networks. While theoretically interesting~\cite{bietti2017invariance}, CKNs perform at lower accuracy than contemporaries, resulting in limited usage. Interestingly, a byproduct of the derivation is a standard Gaussian filter; however, no guidance is provided on its proper integration with existing network components. Instead, we demonstrate practical integration with any strided layer, and empirically show performance increases on a challenging benchmark -- ImageNet classification -- on widely-used networks.

\vspace{-2mm}

\section{Methods}
\label{sec:methods}

\subsection{Preliminaries}
\label{sec:preliminaries}
\vspace{-1mm}

\noindent {\bf Deep convolutional networks as feature extractors} Let an image with resolution $H\times W$ be represented by $X \in \mathds{R}^{H\times W\times 3}$. An $L$-layer CNN can be expressed as a feature extractor $\mathcal{F}_l (X) \in \mathds{R}^{H_l \times W_l \times C_l}$, with layer $l \in \{0, 1, ..., L\}$, spatial resolution $H_l \times W_l$ and $C_l$ channels. Each feature map can also be upsampled to original resolution, $\widetilde{\mathcal{F}}_l (X) \in \mathds{R}^{H \times W \times C_l}$.

\vspace{-1mm}
\noindent {\bf Shift-equivariance and invariance} A function $\mathcal{\widetilde{F}}$ is shift-equivariant if shifting the input equally shifts the output, meaning shifting and feature extraction are commutable.

\vspace{-4.5mm}
\begin{equation}
\Shift _{\Delta h,\Delta w}(\widetilde{\mathcal{F}}(X)) = \widetilde{\mathcal{F}}(\Shift _{\Delta h,\Delta w}(X)) \hspace{4mm} \forall \hspace{1mm} (\Delta h,\Delta w)
\label{eqn:shift-eq}
\end{equation}

A representation is shift-invariant if shifting the input results in an \textit{identical} representation.

\vspace{-5mm}
\begin{equation}
\widetilde{\mathcal{F}}(X) = \widetilde{\mathcal{F}}(\Shift _{\Delta h, \Delta w}(X)) \hspace{4mm} \forall \hspace{1mm} (\Delta h,\Delta w)
\label{eqn:shift-inv}
\end{equation}

\noindent {\bf Periodic-N shift-equivariance/invariance} In some cases, the definitions in Eqns.~\ref{eqn:shift-eq},~\ref{eqn:shift-inv} may hold only when shifts $(\Delta h, \Delta w)$ are integer multiples of N. We refer to such scenarios as periodic shift-equivariance/invariance. For example, periodic-2 shift-invariance means that even-pixel shifts produce an identical output, but odd-pixel shifts may not.

\noindent {\bf Circular convolution and shifting} Edge artifacts are an important consideration. When shifting, information is lost on one side and has to be filled in on the other.

In our CIFAR10 classification experiments, we use circular shifting and convolution. When the convolutional kernel hits the edge, it ``rolls'' to the other side. Similarly, when shifting, pixels are rolled off one edge to the other.

\vspace{-6mm}
\begin{equation}
\begin{split}
[\Shift _{\Delta h,\Delta w}(X)]&_{h,w,c} = X_{(h-\Delta h) \text{\%} H, (w-\Delta w) \text{\%} W,c} \hspace{1mm}, \hspace{1mm} \\
& \text{where \% is the modulus function}
\end{split}
\label{eqn:shift}
\end{equation}
\vspace{-5mm}

The modification minorly affects performance and could be potentially mitigated by additional padding, at the expense of memory and computation. But importantly, this affords us a clean testbed. Any loss in shift-equivariance is purely due to characteristics of the feature extractor.

An alternative is to take a shifted crop from a larger image. We use this approach for ImageNet experiments, as it more closely matches standard train and test procedures.

\subsection{Anti-aliasing to improve shift-equivariance}
\label{sec:our_method}

Conventional methods for reducing spatial resolution -- max-pooling, average pooling, and strided convolution -- all break shift-equivariance. We propose improvements, shown in Figure~\ref{fig:antialias-mods}. We start by analyzing max-pooling.

\noindent {\bf MaxPool$\rightarrow$MaxBlurPool} Consider the example $[0, 0, 1, 1, 0, 0, 1, 1]$ signal in Figure~\ref{fig:simp_ex} (left). Max-pooling (kernel k=$2$, stride s=$2$) will result in $[0, 1, 0, 1]$. Simply shifting the input results in a dramatically different answer of $[1, 1, 1, 1]$. Shift-equivariance is lost. These results are subsampling from an intermediate signal -- the input densely max-pooled (stride-1), which we simply refer to as ``max''. As illustrated in Figure~\ref{fig:fig1} (top), we can write max-pooling as a composition of two functions: $\MaxPool_{k,s} = \Downsample_{s} \circ \Max_{k} $.

The $\Max$ operation preserves shift-equivariance, as it is densely evaluated in a sliding window fashion, but subsequent subsampling does not. We simply propose to add an anti-aliasing filter with kernel $m\times m$, denoted as $\Blur_m$, as shown in Figure~\ref{fig:simp_ex} (right). During implementation, blurring and subsampling are combined, as commonplace in image processing. We call this function $\BlurDown_{m,s}$.

\vspace{-6mm}
\begin{equation}
	\begin{split}
	\MaxPool_{k,s} \rightarrow \hspace{1mm} \Downsample_{s} \circ & \hspace{1mm} \Blur_{m} \circ \Max_{k} \\
				   			 = \BlurDown_{m,s} & \circ \Max_{k}
	\label{eqn:mpblurds}
	\end{split}
\end{equation}
\vspace{-4mm}

Sampling after low-pass filtering gives $[.5, 1, .5, 1]$ and $[.75, .75, .75, .75]$. These are closer to each other and better representations of the intermediate signal.

\noindent {\bf StridedConv$\rightarrow$ConvBlurPool} Strided-convolutions suffer from the same issue, and the same method applies.

\vspace{-8mm}
\begin{equation}
	\Relu \circ \Conv_{k,s} \rightarrow \BlurDown_{m,s} \circ \Relu \circ \Conv_{k,1}
	\label{eqn:mpconv}
	\vspace{-2mm}
\end{equation}

Importantly, this analogous modification applies conceptually to any strided layer, meaning the network designer can keep their original operation of choice.

\begin{figure*}[t!]
    \centering
    \begin{subfigure}
        \centering
        \includegraphics[width=.96\linewidth]{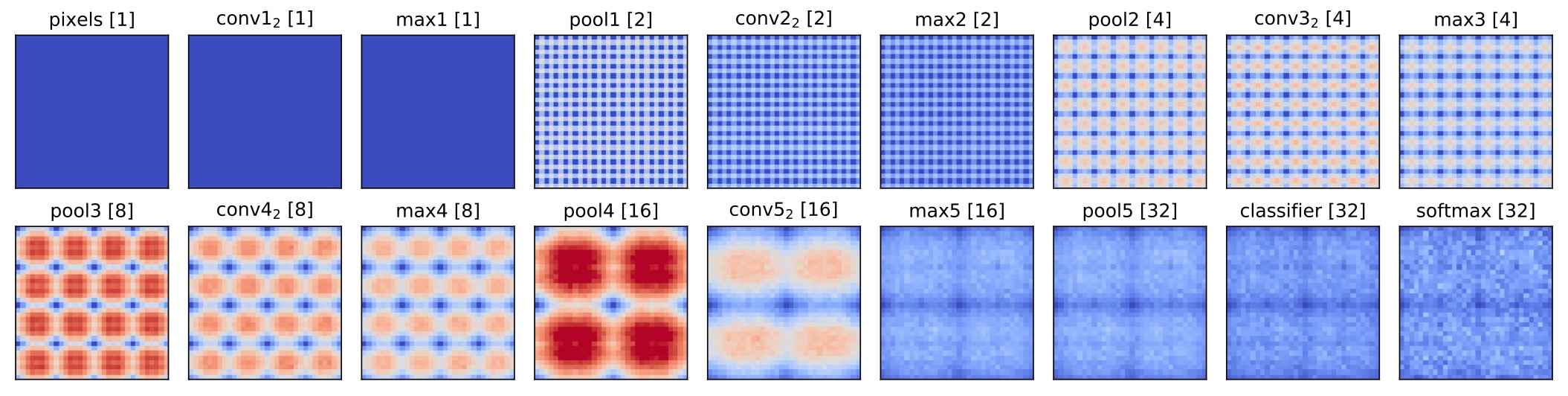}
        \includegraphics[width=.035\linewidth]{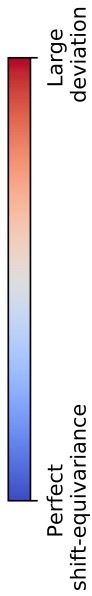} \\
	{\vspace{-2.5mm} (a) Baseline VGG13bn (using MaxPool)}		
    \end{subfigure}
    \vspace{1.5mm}
    \begin{subfigure}
        \centering
        \includegraphics[width=.96\linewidth]{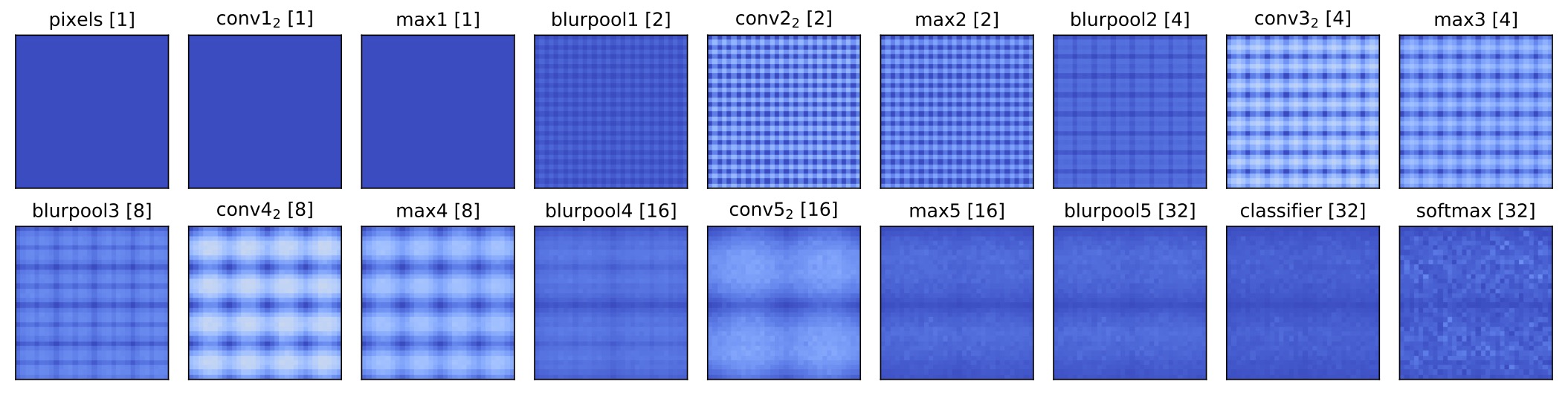}
        \includegraphics[width=.035\linewidth]{figures/sims_shifts_bar.jpg} \\
	{\vspace{-2.5mm} (b) Anti-aliased VGG13bn (using MaxBlurPool, \textit{Bin-5})}		
    \end{subfigure}
    \vspace{-5mm}
    \caption{ {\bf Deviation from perfect shift-equivariance, throughout VGG.}
	Feature distance between left \& right-hand sides of the shift-equivariance condition (Eqn~\ref{eqn:shift-eq}). Each pixel in each heatmap is a shift $(\Delta h, \Delta w)$. \textcolor{darkblue}{Blue} indicates perfect shift-equivariance; \textcolor{darkred}{red} indicates large deviation. Note that the dynamic ranges of distances are different per layer. For visualization, we calibrate by calculating the mean distance between two different images, and mapping \textcolor{darkred}{red} to half the value. Accumulated downsampling factor is in [brackets]; in layers \texttt{pool5}, \texttt{classifier}, and \texttt{softmax}, shift-equivariance and shift-invariance are equivalent, as features have no spatial extent. Layers up to \texttt{max1} have perfect equivariance, as no downsampling yet occurs. {\bf (a)} On the {\bf baseline network}, shift-equivariance is reduced each time downsampling takes place. Periodic-N shift-equivariance holds, with N doubling with each downsampling.
	{\bf (b)} With our {\bf antialiased network}, shift-equivariance is better maintained, and the resulting output is more shift-invariant.
	\vspace{-3mm}
	}
\label{fig:shift-eq-net}
\end{figure*}

\noindent {\bf AveragePool$\rightarrow$BlurPool} Blurred downsampling with a box filter is the same as average pooling. Replacing it with a stronger filter provides better shift-equivariance. We examine such filters next.

\vspace{-3mm}
\begin{equation}
\AvgPool_{k,s} \rightarrow \BlurDown_{m,s}	
\end{equation}

\noindent {\bf Anti-aliasing filter selection} The method allows for a choice of blur kernel. We test $m \times m$ filters ranging from size 2 to 5, with increasing smoothing. The weights are normalized. The filters are the outer product of the following vectors with themselves.

\begin{itemize}[noitemsep,nolistsep,leftmargin=2.5mm] 
\item \textit{\textbf{Rectangle-2}} [1, 1]: moving average or box filter; equivalent to average pooling or ``nearest'' downsampling

\item \textit{\textbf{Triangle-3}} [1, 2, 1]: two box filters convolved together; equivalent to bilinear downsampling

\item \textit{\textbf{Binomial-5}} [1, 4, 6, 4, 1]: the box filter convolved with itself repeatedly; the standard filter used in Laplacian pyramids~\citep{burt1987laplacian}

\end{itemize}

\section{Experiments}
\label{sec:experiments}

\subsection{Testbeds}
\label{sec:testbeds}


\noindent {\bf CIFAR Classification} To begin, we test classification of low-resolution $32\times 32$ images. The dataset contains 50k training and 10k validation images, classified into one of 10 categories. We dissect the VGG architecture~\cite{simonyan2014very}, showing that shift-equivariance is a signal-processing property, progressively lost in each downsampling layer.

\noindent {\bf ImageNet Classification} We then test on large-scale classification on $224\times 224$ resolution images. The dataset contains 1.2M training and 50k validation images, classified into one of 1000 categories. We test across different architecture families -- AlexNet~\cite{krizhevsky2009learning}, VGG~\cite{simonyan2014very}, ResNet~\cite{He_2016_CVPR}, DenseNet~\cite{huang2017densely}, and MobileNet-v2~\cite{sandler2018mobilenetv2} -- with different downsampling strategies, as described in Table~\ref{tab:testbed}. Furthermore, we test the classifier robustness using the Imagenet-C and ImageNet-P datasets~\cite{hendrycks2019using}.

\noindent{\bf Conditional Image Generation} Finally, we show that the same aliasing issues in classification networks are also present in conditional image generation networks. We test on the Labels$\rightarrow$Facades~\cite{tylevcek2013spatial, isola2017image} dataset, where a network is tasked to generated a 256$\times$256 photorealistic image from a label map. There are 400 training and 100 validation images.

\subsection{Shift-Invariance/Equivariance Metrics}

Ideally, a shift in the input would result in equally shifted feature maps internally:

\noindent \textbf{\bf Internal feature distance.} We examine internal feature maps with $d(\Shift _{\Delta h,\Delta w}(\widetilde{\mathcal{F}}(X)),\widetilde{\mathcal{F}}(\Shift _{\Delta h,\Delta w}(X)))$ (left \& right-hand sides of Eqn.~\ref{eqn:shift-eq}). We use cosine distance, as common for deep features~\citep{kiros2015skip,zhang2018unreasonable}.

We can also measure the stability of the output: 

\noindent \textbf{\bf Classification consistency.} For classification, we check how often the network outputs the same classification, given the same image with two different shifts: $\mathds{E}_{X, h_1, w_1, h_2, w_2} \mathds{1}\{\arg\max P(\Shift _{h_1, w_1}(X))=\arg\max P(\Shift _{h_2, w_2}(X))\}$.

\noindent \textbf{\bf Generation stability.} For image translation, we test if a shift in the input image generates a correspondingly shifted output. For simplicity, we test horizontal shifts. $\mathds{E}_{X, \Delta w} \text{PSNR} \big( \Shift_{0, \Delta w} (\mathcal{F}(X)) ),  \mathcal{F}( \Shift_{0, \Delta w} (X) ) \big)$.

\begin{table}
\scalebox{.75} {
 \begin{tabular}{c c c c c c c}
 \toprule
 & \multicolumn{5}{c}{\bf ImageNet Classification} & {\bf Generation} \\
\cmidrule(lr){2-6} \cmidrule(lr){7-7}
& {\bf Alex-} & {\bf VGG} & {\bf Res-} & {\bf Dense-} & {\bf Mobile-} & {\bf U-} \\
& {\bf Net} & {\bf } & {\bf Net} & {\bf Net} & {\bf Netv2} & {\bf Net}\\
 \cmidrule(lr){1-7} 
{\bf StridedConv} & 1$^{\diamond}$ & -- & $4^{\ddagger}$ & $1^{\ddagger}$ & $5^{\ddagger}$ & 8 \\
{\bf MaxPool} & 3 & 5 & 1 & 1 & -- & -- \\
{\bf AvgPool} & -- & -- & -- & 3 & -- & -- \\ 
\bottomrule
\vspace{-11mm}
\end{tabular}
}
\caption{{\bf Testbeds.} We test across tasks (ImageNet classification and Labels$\rightarrow$Facades) and network architectures. Each architecture employs different downsampling strategies. We list how often each is used here. We can antialias each variant. $^{\diamond}$This convolution uses stride 4 (all others use 2). We only apply the antialiasing at stride 2. Evaluating the convolution at stride 1 would require large computation at full-resolution. $^{\ddagger}$For the same reason, we do not antialias the first strided-convolution in these networks.
\label{tab:testbed}
}
\end{table}

\subsection{Internal shift-equivariance}

We first test on the CIFAR dataset using the VGG13-bn~\cite{simonyan2014very} architecture.

We dissect the progressive loss of shift-equivariance by investigating the VGG architecture internally. The network contains 5 blocks of convolutions, each followed by max-pooling (with stride 2), followed by a linear classifier. For purposes of our understanding, MaxPool layers are broken into two components -- before and after subsampling, e.g., \texttt{max1} and \texttt{pool1}, respectively. In Figure~\ref{fig:shift-eq-net} (top), we show internal feature distance, as a function of all possible shift-offsets $(\Delta h, \Delta w)$ and layers. All layers before the first downsampling, \texttt{max1}, are shift-equivariant. Once downsampling occurs in \texttt{pool1}, shift-equivariance is lost. However, periodic-N shift-equivariance still holds, as indicated by the stippling pattern in \texttt{pool1}, and each subsequent subsampling doubles the factor N.

In Figure~\ref{fig:shift-eq-net} (bottom), we plot shift-equivariance maps with our anti-aliased network, using MaxBlurPool. Shift-equivariance is clearly better preserved. In particular, the severe drop-offs in downsampling layers do not occur. Improved shift-equivariance throughout the network cascades into more consistent classifications in the output, as shown by some selected examples in Figure~\ref{fig:class_stab}. This study uses a \textit{Bin-5} filter, trained without data augmentation. The trend holds for other filters and when training with augmentation.


\subsection{Large-scale ImageNet classification}

\subsubsection{Shift-invariance and accuracy}

\begin{figure}[t]
\centering
\includegraphics[width=1.\linewidth, trim={.2cm .2cm 0 .2cm}, clip]{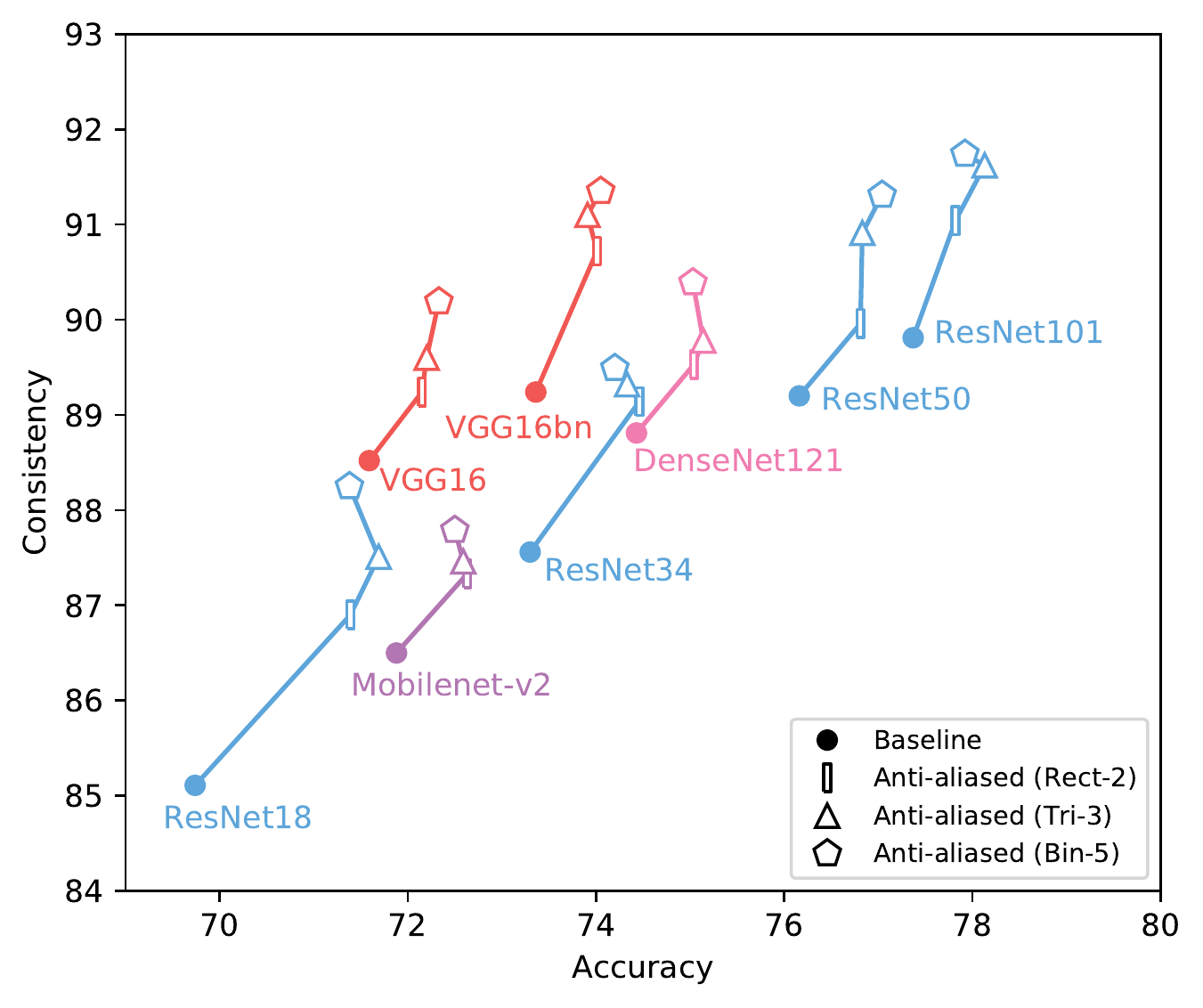}
\vspace{-10mm}
\caption{\textbf{ImageNet Classification consistency vs. accuracy.} Up (more consistent to shifts) and to the right (more accurate) is better. Different shapes correspond to the baseline (circle) or variants of our anti-aliased networks (bar, triangle, pentagon for length 2, 3, 5 filters, respectively). We test across network architectures. As expected, low-pass filtering helps shift-invariance. Surprisingly, classification accuracy is also improved.
\label{fig:consist_vs_acc_imagenet_agg}
\vspace{-2mm}
}
\end{figure}

We next test on large-scale image classification of ImageNet~\cite{russakovsky2015imagenet}. In Figure~\ref{fig:consist_vs_acc_imagenet_agg}, we show classification accuracy and consistency, across variants of several architectures -- VGG, ResNet, DenseNet, and MobileNet-v2. The off-the-shelf networks are labeled as \textit{Baseline}, and we use standard training schedules from the publicly available PyTorch~\cite{paszke2017automatic} repository for our anti-aliased networks. Each architecture has a different downsampling strategy, shown in Table~\ref{tab:testbed}.
We typically refer to the popular ResNet50 as a running example; note that we see similar trends across network architectures.

\noindent {\bf Improved shift-invariance} We apply progressively stronger filters -- \textit{Rect-2}, \textit{Tri-3}, \textit{Bin-5}. Doing so increases ResNet50 stability by $+0.8\%$, $+1.7\%$, and $+2.1\%$, respectively. Note that doubling layers -- going to ResNet101 -- only increases stability by $+0.6\%$. Even a simple, small low-pass filter, directly applied to ResNet50, outpaces this. As intended, stability increases across architectures (points move upwards in Figure~\ref{fig:consist_vs_acc_imagenet_agg}).

\noindent {\bf Improved classification} Filtering improves the shift-invariance. How does it affect absolute classification performance? We find that across the board, \textit{performance actually increases} (points move to the right in Figure~\ref{fig:consist_vs_acc_imagenet_agg}). The filters improve ResNet50 by $+0.7\%$ to $+0.9\%$. For reference, doubling the layers to ResNet101 increases accuracy by $+1.2\%$. A low-pass filter makes up much of this ground, without adding any learnable parameters. This is a surprising, unexpected result, as low-pass filtering removes information, and could be expected to reduce performance. On the contrary, we find that it serves as effective regularization, and these widely-used methods improve with simple anti-aliasing. As ImageNet-trained nets often serve as the backbone for downstream tuning, this improvement may be observed across other applications as well.

The best performing filter varies by architecture, but all filters improve over the baseline. We recommend using the \textit{Tri-3} or \textit{Bin-5} filter. If shift-invariance is especially desired, stronger filters can be used.

\subsubsection{Out-of-distribution Robustness}

We have shown increased stability (to shifts), as well as accuracy. Next, we test the generalization capability the classifier in these two aspects, using datasets from~\citet{hendrycks2019using}. We test stability to perturbations other than shifts. We then test accuracy on systematically corrupted images. Results are shown in Table~\ref{tab:results-imagenetcp}, averaged across corruption types. We show the raw, unnormalized average, along with a weighted ``normalized'' average, as recommended.

\noindent {\bf Stability to perturbations} The ImageNet-P dataset~\cite{hendrycks2019using} contains short video clips of a single image with small perturbations added, such as variants of noise (Gaussian and shot), blur (motion and zoom), simulated weather (snow and brightness), and geometric changes (rotation, scaling, and tilt). Stability is measured by flip rate (mFR) -- how often the top-1 classification changes,
on average, in consecutive frames. Baseline ResNet50 flips $7.9\%$ of the time; adding anti-aliasing \textit{Bin-5} reduces by $1.0\%$. While antialiasing provides increased stability to shifts by design, a ``free'', emergent property is increased stability to other perturbation types.

\noindent {\bf Robustness to corruptions} We observed increased accuracy on clean ImageNet. Here, we also observe more graceful degradation when images are corrupted. In addition to the previously explored corruptions, ImageNet-C contains impulse noise, defocus and glass blur, simulated frost and fog, and various digital alterations of contrast, elastic transformation, pixelation and jpeg compression. The geometric perturbations are not used. ResNet50 has mean error rate of $60.6\%$. Anti-aliasing with \textit{Bin-5} reduces the error rate by $2.5\%$. As expected, the more ``high-frequency'' corruptions, such as adding noise and pixelation, show greater improvement. Interestingly, we see improvements even with ``low-frequency'' corruptions, such defocus blur and zoom blur operations as well.

\begin{table}[t]
\scalebox{.88} {
 \begin{tabular}{c c c c c c}
 \toprule
 & \multicolumn{2}{c}{\bf Normalized average} &  \multicolumn{2}{c}{\bf Unnormalized average}  \\ \cmidrule(lr){2-3} \cmidrule(lr){4-6}
 & {\bf ImNet-C} & {\bf ImNet-P} & {\bf ImNet-C} & {\bf ImNet-P} \\ \cmidrule(lr){2-2} \cmidrule(lr){3-3} \cmidrule(lr){4-4} \cmidrule(lr){5-5}
 & {\bf mCE} & {\bf mFR}  & {\bf mCE} & {\bf mFR} \\ \midrule
{\bf Baseline} & 76.4 & 58.0 & 60.6 & 7.92 \\ \cdashline{1-5}
{\bf Rect-2} & 75.2 & 56.3 & 59.5 & 7.71 \\
{\bf Tri-3} & 73.7 & 51.9 & 58.4 & 7.05 \\
{\bf Bin-5} & {\bf 73.4} & {\bf 51.2} & {\bf 58.1} & {\bf 6.90} \\
\bottomrule
\vspace{-5mm}
\end{tabular}
}
\caption{{\bf Accuracy and stability robustness.} Accuracy in ImageNet-C, which contains systematically corrupted ImageNet images, measured by mean corruption error {\bf mCE} (lower is better). Stability on ImageNet-P, which contains perturbed image sequences, measured by mean flip rate {\bf mFR} (lower is better). We show raw, unnormalized scores, as well as scores normalized to AlexNet, as used in~\citet{hendrycks2019using}. Anti-aliasing improves both accuracy and stability over the baseline. All networks are variants of ResNet50.
\vspace{-3mm}
\label{tab:results-imagenetcp}
}
\vspace{-2mm}
\end{table}

%

Together, these results indicate that a byproduct of antialiasing is a more robust, generalizable network. Though motivated by shift-invariance, we actually observe increased stability to other perturbation types, as well as increased accuracy, both on clean and corrupted images.

\begin{figure*}[t]
\centering
\includegraphics[width=1.\linewidth]{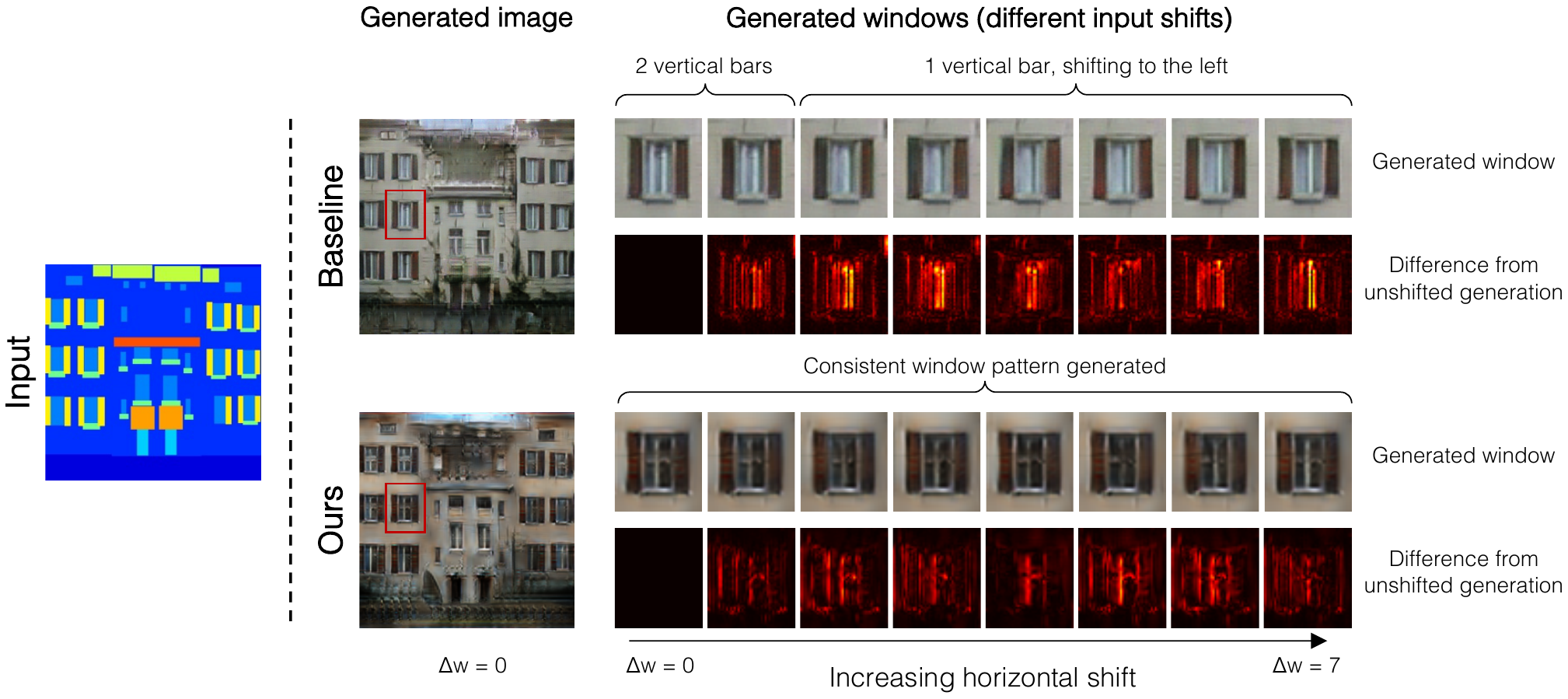}
\vspace{-10mm}
\caption{
{\bf Selected example of generation instability.} The left two images are generated facades from label maps. For the baseline method (top), input shifts cause different window patterns to emerge, due to naive downsampling and upsampling. Our method (bottom) stabilizes the output, generating the same window pattern, regardless the input shift.
\label{fig:gen_stab}
\vspace{-4mm}
}
\end{figure*}

\subsection{Conditional image generation (Label$\rightarrow$Facades)}

We test on image generation, outputting an image of a facade given its semantic label map~\cite{tylevcek2013spatial}, in a GAN setup~\cite{goodfellow2014generative, isola2017image}. Our classification experiments indicate that anti-aliasing is a natural choice for the discriminator, and is used in the recent StyleGAN method~\cite{karras2018style}. Here, we explore its use in the generator, for the purposes of obtaining a shift-equivariant image-to-image translation network.

\noindent \textbf{Baseline} We use the pix2pix method~\cite{isola2017image}. The method uses U-Net~\cite{ronneberger2015u}, which contains 8 downsampling and 8 upsampling layers, with skip connections to preserve local information. No anti-aliasing filtering is applied in down or upsampling layers in the baseline. In Figure~\ref{fig:gen_stab}, we show a qualitative example, focusing in on a specific window. In the baseline (top), as the input $X$ shifts horizontally by $\Delta w$, the vertical bars on the generated window also shift. The generations start with two bars, to a single bar, and eventually oscillates back to two bars.
A shift-equivariant network would provide the same resulting facade, no matter the shift.

\begin{table}
\scalebox{.85}{
    \begin{tabular}{c c c c c c}
    \toprule
	 & {\bf Baseline} & {\bf Rect-2} & {\bf Tri-3} & {\bf Bin-4} & {\bf Bin-5} \\ \hline
      Stability [dB] & \multirow{1}{*}{29.0} & \multirow{1}{*}{30.1} & \multirow{1}{*}{30.8} & \multirow{1}{*}{31.2} & \multirow{1}{*}{34.4} \\
      TV Norm $\times100$ & 7.48 & 7.07 & 6.25 & 5.84 & 6.28 \\ \bottomrule
    \end{tabular}
    \vspace{-4mm}
	}
    \vspace{-4mm}
    \caption{{\bf Generation stability} PSNR (higher is better) between generated facades, given two horizontally shifted inputs. More aggressive filtering in the down and upsampling layers leads to a more shift-equivariant generator. {\bf Total variation (TV) of generated images} (closer to ground truth images 7.80 is better). Increased filtering decreases the frequency content of generated images.
    \vspace{-5mm}
    \label{tab:gen_stab}
    }
\end{table}

\noindent \textbf{Applying anti-aliasing} We augment the strided-convolution downsampling by blurring. The U-Net also uses upsampling layers, without any smoothing. Similar to the subsampling case, this leads to aliasing, in the form of grid artifacts~\cite{odena2016deconvolution}. We mirror the downsampling by applying the same filter after upsampling. Note that applying the \textit{Rect-2} and \textit{Tri-3} filters while upsampling correspond to ``nearest'' and ``bilinear'' upsampling, respectively. By using the \textit{Tri-3} filter, the same window pattern is generated, regardless of input shift, as seen in Figure~\ref{fig:gen_stab} (bottom). 

We measure similarity using peak signal-to-noise ratio between generated facades with shifted and non-shifted inputs: $\mathds{E}_{X, \Delta w} \text{PSNR}( \Shift_{0,\Delta w}(F(X)), F(\Shift _{0,\Delta w}(X))))$. In Table~\ref{tab:gen_stab}, we show that the smoother the filter, the more shift-equivariant the output.

A concern with adding low-pass filtering is the loss of ability to generate high-frequency content, which is critical for generating high-quality imagery. Quantitatively, in Table~\ref{tab:gen_stab}, we compute the total variation (TV) norm of the generated images. Qualitatively, we observe that generation quality typically holds with the \textit{Tri-3} filter and subsequently degrades. In the supplemental material, we show examples of applying increasingly aggressive filters. We observe a boost in shift-equivariance while maintaining generation quality, and then a tradeoff between the two factors.

These experiments demonstrate that the technique can make a drastically different architecture (U-Net) for a different task (generating pixels) more shift-equivariant.

\vspace{-1.5mm}
\section{Conclusions and Discussion}
\label{sec:conclusions}

Shift-equivariance is lost in modern deep networks, as commonly used downsampling layers ignore Nyquist sampling and alias. We integrate low-pass filtering to anti-alias, a common signal processing technique. The simple modification achieves higher consistency, across architectures and downsampling techniques. In addition, in classification, we observe surprising boosts in accuracy and robustness.

Anti-aliasing for shift-equivariance is well-understood. A future direction is to better understand how it affects and improves generalization, as we observed empirically.
Other directions include the potential benefit to downstream applications, such as nearest-neighbor retrieval, improving temporal consistency in video models, robustness to adversarial examples, and high-level vision tasks such as detection. Adding the inductive bias of shift-invariance serves as ``built-in'' shift-based data augmentation. This is potentially applicable to online learning scenarios, 
where the data distribution is changing.

\subsubsection*{Acknowledgments}

I am especially grateful to Eli Shechtman for helpful discussion and guidance. Micha\"{e}l Gharbi, Andrew Owens, and anonymous reviewers provided beneficial feedback on earlier drafts. I thank labmates and mentors, past and present -- Sylvain Paris, Oliver Wang, Alexei A. Efros, Angjoo Kanazawa, Taesung Park, and Phillip Isola -- for their helpful comments and encouragement.  I thank Dan Hendrycks for discussion about robustness tests on ImageNet-C/P.

\subsubsection*{Changelog}

{\bf v1} ArXiv preprint. Paper accepted to ICML 2019.

{\bf v2} ICML camera ready. Added additional networks. Added robustness measures. ImageNet consistency numbers and AlexNet results re-evaluated; small fluctuations but no changes in general trends. Compressed main paper to 8 pages. Cifar results moved to supplemental. Small changes to text.

\bibliography{camera-ready}
\bibliographystyle{icml2019}


\appendix

\section*{Supplementary Material}

Here, we show additional results and experiments for CIFAR classification, ImageNet expanded results, and conditional image generation.

\section{CIFAR Classification}

\subsection{Classification results}

\begin{figure*}
\centering
{\hspace{8mm} VGG-13~\cite{simonyan2014very} \hspace{24mm} DenseNet-40-12~\cite{huang2017densely}}

\includegraphics[width=1.\linewidth, trim={0cm .2cm 0 .2cm}, clip]{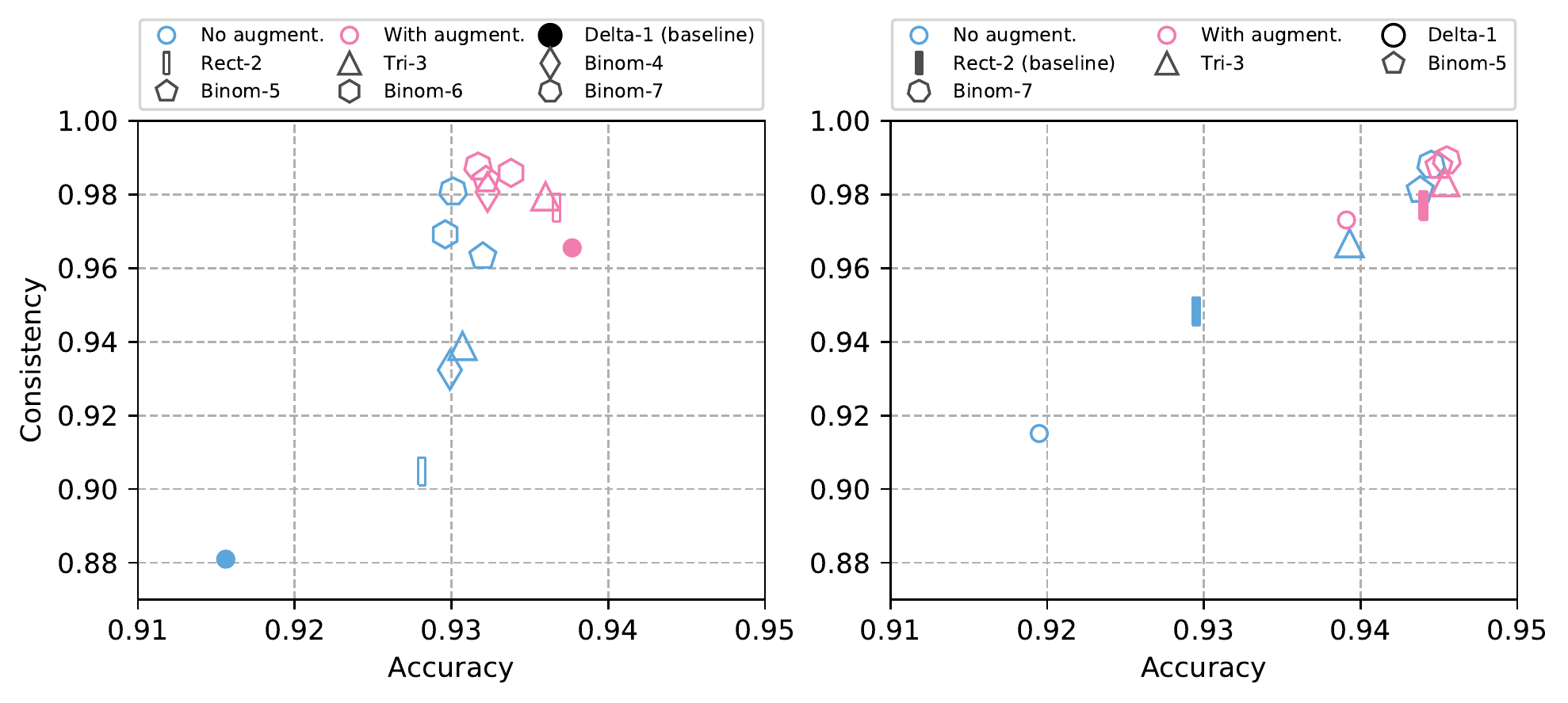}

\vspace{-3mm}
\caption{\textbf{CIFAR10 Classification consistency vs. accuracy.}
VGG {\bf (left)} and DenseNet {\bf (right)} networks.
Up (more consistent) and to the right (more accurate) is better. Number of sides corresponds to number of filter taps used (e.g., diamond for 4-tap filter); colors correspond to filters trained without (blue) and with (pink) shift-based data augmentation, using various filters. We show accuracy for no shift when training without shifts, and a random shift when training with shifts.
\label{fig:consist_vs_acc}
\vspace{-2mm}
}
\end{figure*}

We train both without and with shift-based data augmentation. We evaluate on classification accuracy and consistency. The results are shown in Table~\ref{tab:results-cifar-supp} and Figure~\ref{fig:consist_vs_acc}.

In the main paper, we showed internal activations on the older VGG~\cite{simonyan2014very} network. Here, we also present classification accuracy and consistency results on the output, along with the more modern DenseNet~\cite{huang2017densely} architecture.

\begin{table}[h]
\scalebox{.9} {
 \begin{tabular}{c c c}
 \toprule
\multirow{2}{*}{\bf Architecture} & \multicolumn{2}{c}{\bf Classification (CIFAR)} \\
\cmidrule(lr){2-3}
& {\bf VGG13-bn} & {\bf DenseNet-40-12} \\ \cmidrule(lr){1-3} 
{\bf StridedConv} & -- & -- \\
{\bf MaxPool} & 5 & -- \\
{\bf AvgPool} & -- & 2 \\ 
\bottomrule
\vspace{-8mm}
\end{tabular}
}
\caption{{\bf Testbeds (CIFAR10 Architectures).} We use slightly different architectures for VGG~\cite{simonyan2014very} and DenseNet~\cite{huang2017densely} than the ImageNet counterparts.
\label{tab:testbed-cifar}
}
\end{table}

\noindent {\bf Training without data augmentation} Without the benefit of seeing shifts at training time, the baseline network produces inconsistent classifications -- random shifts of the same image only agree $88.1\%$ of the time. Our anti-aliased network, with the MaxBlurPool operator, increases consistency. The larger the filter, the more consistent the output classifications. This result agrees with our expectation and theory -- improving shift-equivariance throughout the network should result in more consistent classifications across shifts, even when such shifts are not seen at training.

In this regime, accuracy clearly increases with consistency, as seen with the blue markers in Figure~\ref{fig:consist_vs_acc}. Filtering does not destroy the signal or make learning harder. On the contrary, shift-equivariance serves as ``built-in'' augmentation, indicating more efficient data usage.

\noindent {\bf Training with data augmentation} In principle, networks can \textit{learn} to be shift-invariant from data. Is data augmentation all that is needed to achieve shift-invariance? By applying the \textit{Rect-2} filter, a large increase in consistency, $96.6\rightarrow 97.6$, can be had at a small decrease in accuracy $93.8\rightarrow 93.7$. Even when seeing shifts at training, antialiasing increases consistency. From there, stronger filters can increase consistency, at the expense of accuracy.

\begin{table*}
\scalebox{.9} {
 \begin{tabular}{c c c c c c c c c c}
 \toprule

\multirow{4}{*}{\bf Net} & \multirow{4}{*}{\bf Filter shape} & \multirow{4}{*}{\bf \# Taps} & \multirow{4}{*}{\bf Weights} & \multicolumn{3}{c}{\bf Train w/o augmentation} & \multicolumn{3}{c}{\bf Train w/ augmentation} \\ \cmidrule(lr){5-7} \cmidrule(lr){8-10}

 & &  &  & \multicolumn{2}{c}{\bf Accuracy} & \multirow{2}{*}{\bf Consistency} & \multicolumn{2}{c}{\bf Accuracy} & \multirow{2}{*}{\bf Consistency} \\ \cmidrule(lr){5-6} \cmidrule(lr){8-9}

 & &  &  & {\bf None} & {\bf Rand} & & {\bf None} & {\bf Rand} & \\ \midrule

\multirow{6}{*}{\bf VGG} & {\bf Delta (baseline)} & 1 & [1] & \worst{91.6} & \worst{87.4} & \worst{88.1} & 93.4 & \textbf{93.8} & \worst{96.6} \\   \cdashline{2-10}

& {\bf Rectangle} & 2 & [1, 1] & 92.8 & \worst{89.3} & \worst{90.5} & \textbf{93.9} & 93.7 & \worst{97.6} \\

& {\bf Triangle} & 3 & [1, 2, 1] & 93.1 & 91.4 & 93.9 & 93.6 & \best{93.6} & 98.0 \\

& {\bf Binomial} & 4 & [1, 3, 3, 1] & 93.0 & 91.1 & 93.2 & 93.4 & 93.2 & 98.1 \\

& {\bf Binomial} & {5} & {[1, 4, 6, 4, 1]} & \textbf{93.2} & 92.6 & 96.3 & 93.1 & 93.2 & 98.4 \\

& {\bf Binomial} & {6} & {[1, 5, 10, 10, 5, 1]} & 93.0 & 92.4 & 96.9 & 93.4 & 93.4 & 98.6 \\

& {\bf Binomial} & {7} & {[1, 6, 15, 20, 15, 6, 1]} & 93.0 & \textbf{93.0} & \textbf{98.1} & 93.2 & 93.2 & \textbf{98.8} \\   \midrule

\multirow{5}{*}{\bf Dense} & {\bf Delta} & 1 & [1] & 92.0 &	89.9 &	91.5 &	93.9 &	93.9 &	97.3 \\
& {\bf Rect (baseline)} & 2 & [1, 1] & 93.0 &	92.3 &	94.8 &	94.4 &	94.4 &	97.7 \\  \cdashline{2-10}
& {\bf Triangle} & 3 & [1, 2, 1] &	93.9 &	93.5 &	96.7 &	{\bf 94.5} & 	94.5 &	98.3 \\
& {\bf Binomial} & 5 & [1, 4, 6, 4, 1] &	94.4 &	94.0 &	98.1 &	{\bf 94.5} & 	94.5 &	98.8 \\
& {\bf Binomial} & 7 & [1, 6, 15, 20, 15, 6, 1] & {\bf 94.5} &	{\bf 94.3} & {\bf 98.8} &	{\bf 94.5} & {\bf 94.6} & {\bf 98.9} \\

\bottomrule

\vspace{-5mm}
\end{tabular}
}
\vspace{-2mm}
\caption{{\bf CIFAR Classification accuracy and consistency} Results across blurring filters and training scenarios (without and with data augmentation). We evaluate classification accuracy without shifts ({\bf Accuracy -- None}) and on random shifts ({\bf Accuracy -- Random}), as well as classification {\bf consistency}. 
\vspace{-2mm}
\label{tab:results-cifar-supp}
}
\end{table*}

\begin{figure*}[t]
\centering
\includegraphics[height=4.3cm, trim={0 0 0 0}, clip]{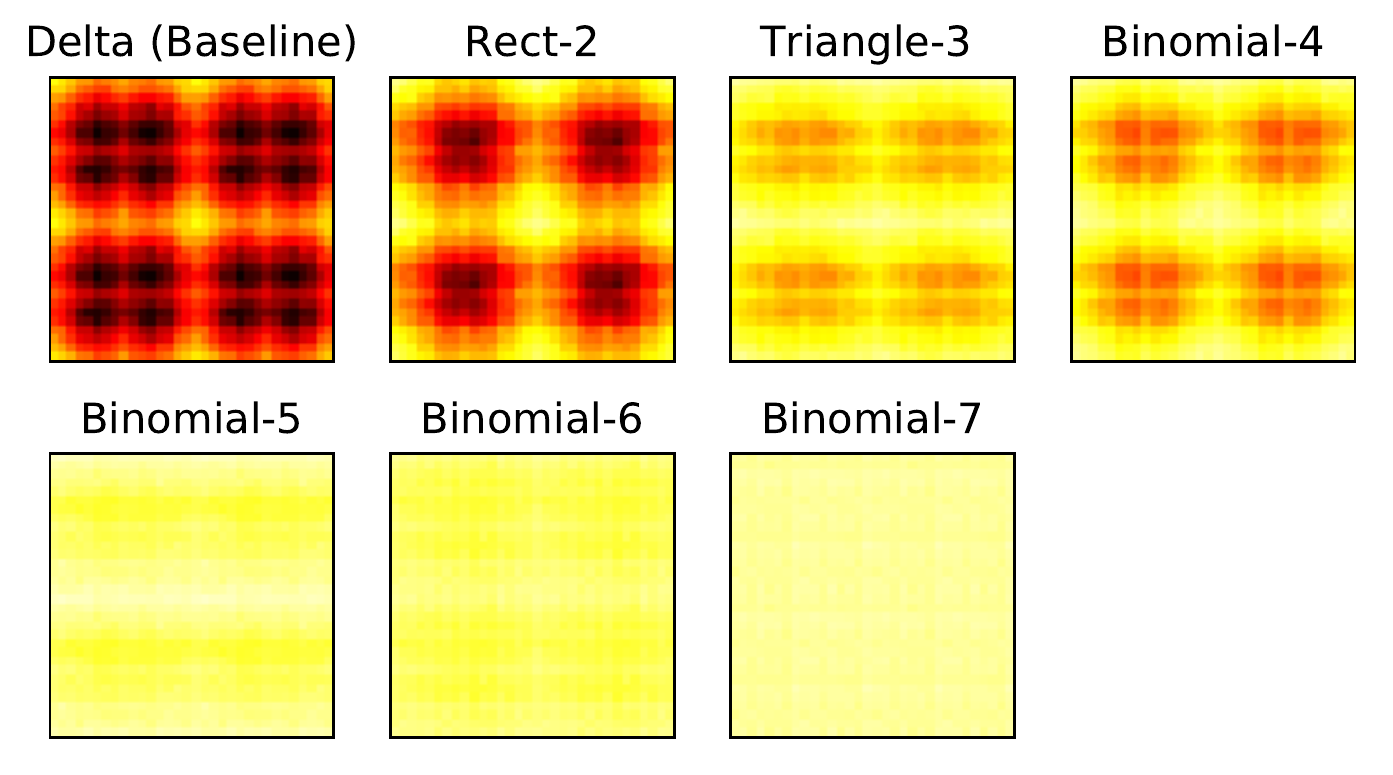}
\includegraphics[height=4.3cm, trim={10.5cm 0 0 0}, clip]{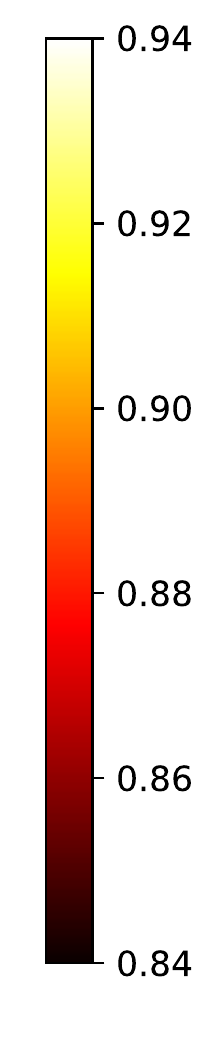}
\includegraphics[height=4.5cm, trim={0.1cm 0 0 0}, clip]{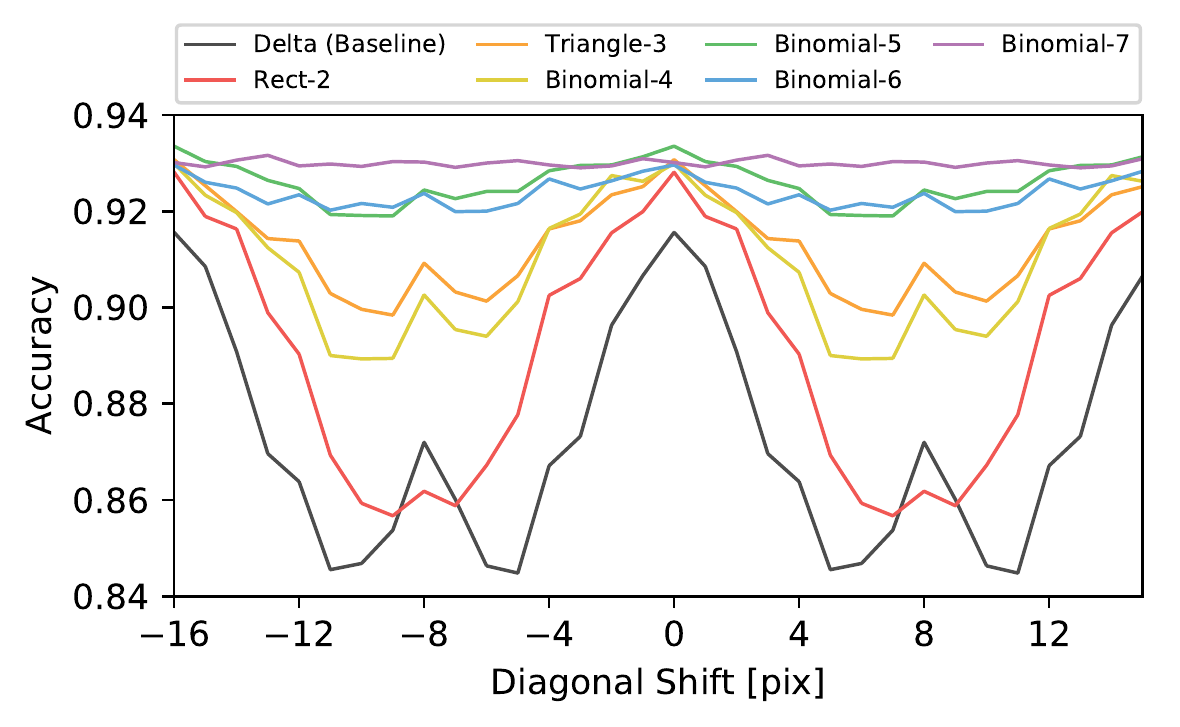}

\caption{\textbf{Average accuracy as a function of shift.} {\bf (Left)} We show classification accuracy across the test set as a function of shift, given different filters. {\bf (Right)} We plot accuracy vs diagonal shift in the input image, across different filters. Note that accuracy degrades quickly with the baseline, but as increased filtering is added, classifications become consistent across spatial positions.
\label{fig:acc_vs_pos}
\vspace{-2mm}
}
\end{figure*}

\begin{figure}[h]
  \begin{center}
	\includegraphics[width=1.\columnwidth,trim={0.2cm 0 0.3cm 0.1cm},clip]{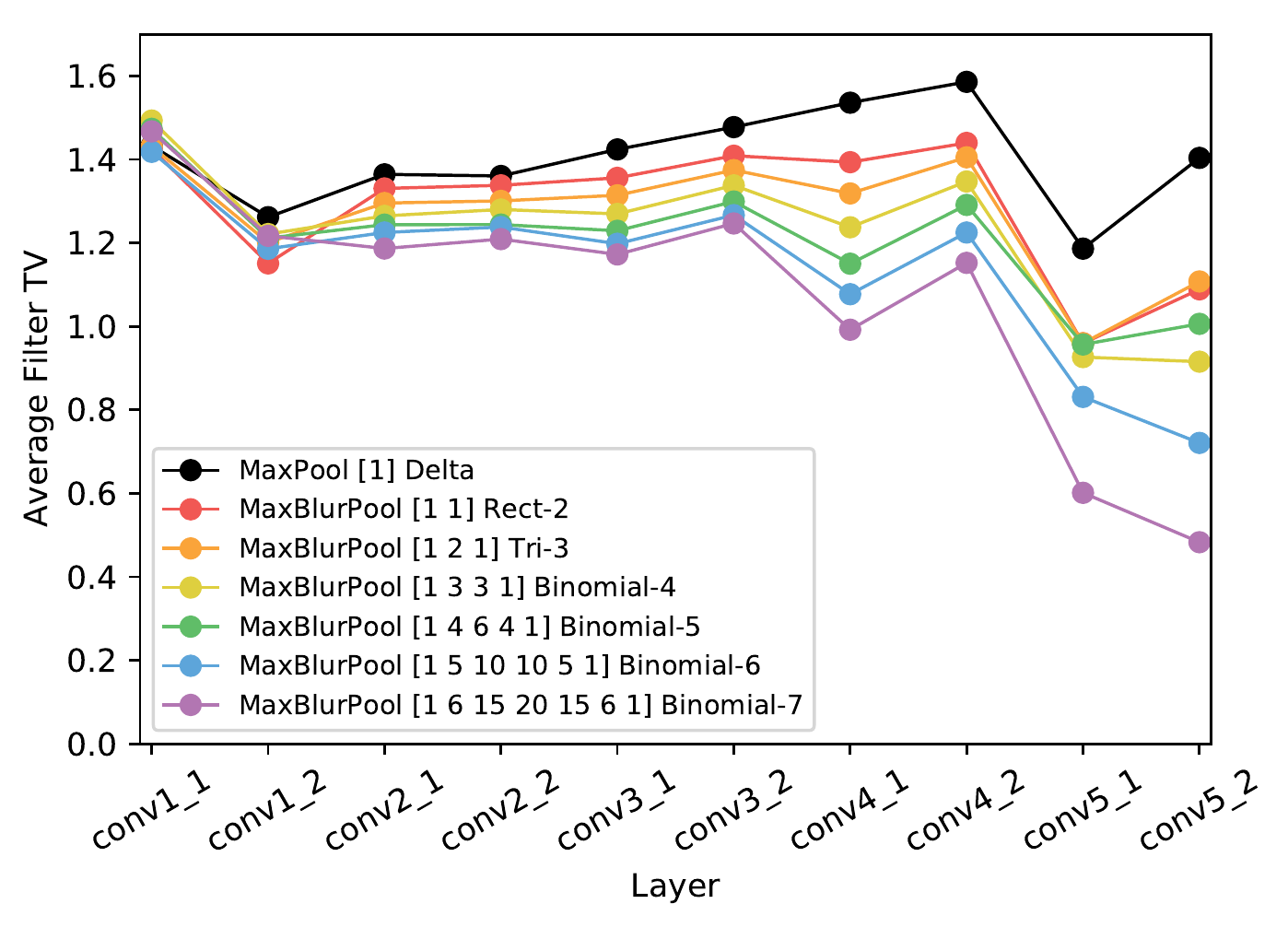}
  \end{center}
  \vspace{-7mm}
  \caption{{\bf Total Variation (TV) by layer.} We compute average smoothness of learned conv filters per layer (lower is smoother). Baseline MaxPool is in black, and adding additional blurring is shown in colors. Note that the \textit{learned} convolutional layers become smoother, indicating that a smoother feature extractor is induced.
  \label{fig:filt_tv}
}
\end{figure}

\begin{figure*}[h]
\centering
\includegraphics[width=1.\linewidth]{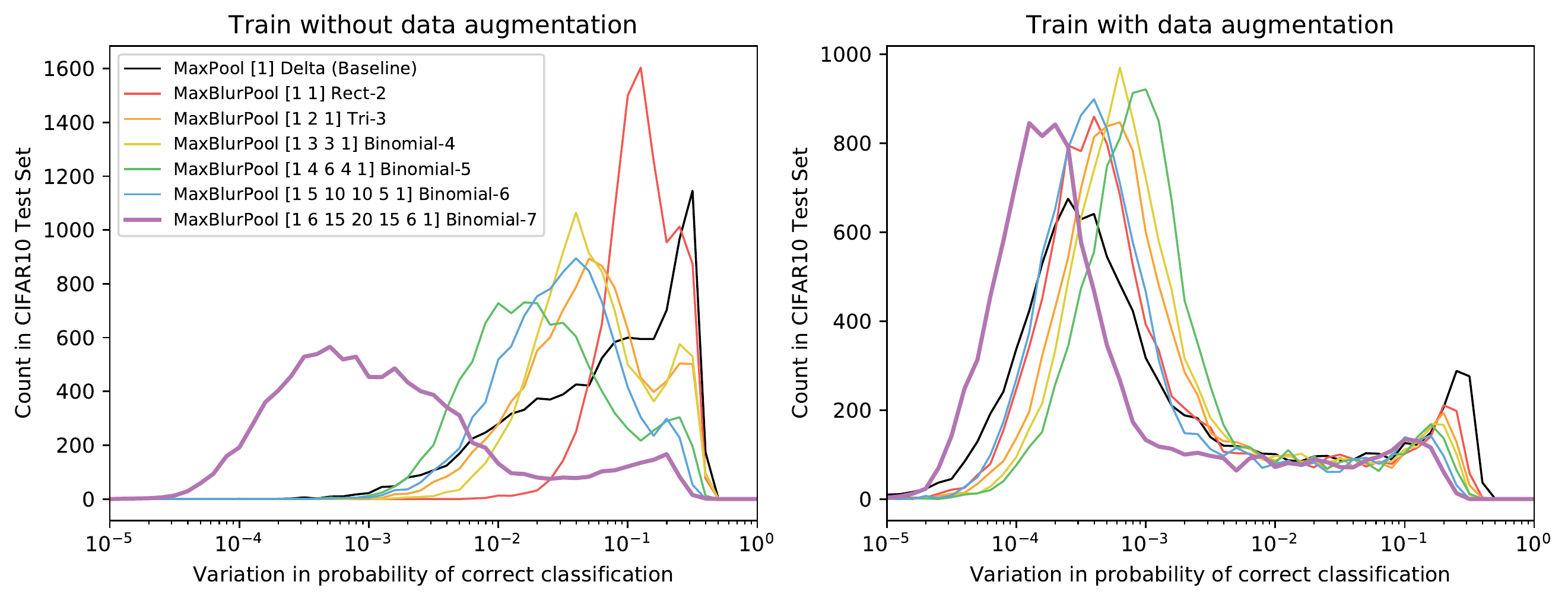}
\vspace{-10mm}
\caption{\textbf{Distribution of per-image classification variation.}
We show the distribution of classification variation in the test set, {\bf (left)} without and {\bf (right)} with data augmentation at training. Lower variation means more consistent classifications (and increased shift-invariance). Training with data augmentation drastically reduces variation in classification. Adding filtering further decreases variation. 
\label{fig:variation_change}
\vspace{-4mm}
}
\end{figure*}

\begin{figure*}[h!]
\centering
{\footnotesize \hspace{4mm} Train without Data Augmentation \hspace{45mm} Train with Data Augmentation}
\includegraphics[width=1.\linewidth, trim={0 .2cm 0 .15cm}, clip]{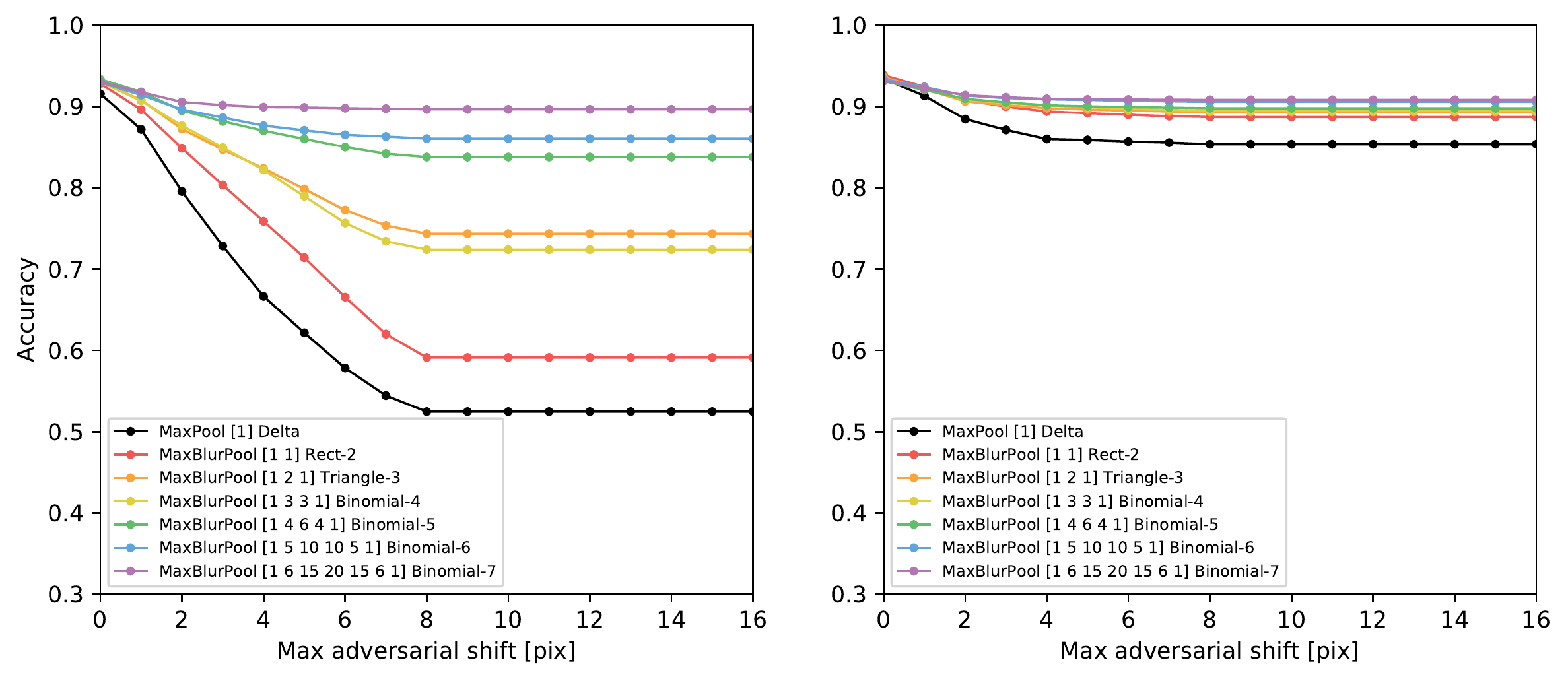}
\vspace{-7mm}
\caption{\textbf{Robustness to shift-based adversarial attack.} Classification accuracy as a function of the number of pixels an adversary is allowed to shift the image. Applying our proposed filtering increases robustness, both without {\bf (left)} and with {\bf {right}} data augmentation.
\label{fig:shift-adv}
}
\end{figure*}

\noindent {\bf DenseNet results} We show a summary of VGG and DenseNet in Table~\ref{tab:testbed-cifar}. DenseNet uses comparatively fewer downsampling layers -- 2 average-pooling layers instead of 5 max-pooling layers. With just two downsampling layers, the baseline still loses shift-invariance. Even when training with data augmentation, replacing average-pooling with blurred-pooling increases both consistency and even minorly improves accuracy. Note that the DenseNet architecture performs stronger than VGG to begin with. In this setting, the \textit{Bin-7} BlurPool operator works best for both consistency and accuracy. Again, applying the operator serves are ``built-in" data augmentation, performing strongly even without shifts at train time.

\noindent {\bf How do the learned convolutional filters change?} Our proposed change smooths the internal feature maps for purposes of downsampling. How does training with this layer affect the \textit{learned} convolutional layers? We measure spatial smoothness using the normalized Total Variation (TV) metric proposed in~\citet{ruderman2018pooling}. 
A higher value indicates a filter with more high-frequency components. A lower value indicates a smoother filter. As shown in Figure~\ref{fig:filt_tv}, the anti-aliased networks (red-purple) actually learn smoother filters throughout the network, relative to the baseline (black). Adding in more aggressive low-pass filtering further decreases the TV (increasing smoothness). This indicates that our method actually induces a smoother feature extractor overall.

\noindent {\bf Timing analysis}
The average speed of a forward pass of VGG13bn using batch size 100 CIFAR images on a GTX1080Ti GPU is 10.19ms. Evaluating $\Max$ at stride 1 instead of 2 adds $3.0\%$. From there, low-pass filtering with kernel sizes 3, 5, 7 adds additional $5.5\%, 7.6\%, 9.3\%$ time, respectively, relative to baseline. The method can be implemented more efficiently by separating the low-pass filter into horizontal and vertical components, allowing added time to scale linearly with filter size, rather than quadratically.
In total, the largest filter adds $12.3\%$ per forward pass. This is significantly cheaper than evaluating multiple forward passes in an ensembling approach ($1024\times$ computation to evaluate every shift), or evaluating each layer more densely by exchanging striding for dilation ($4\times, 16\times, 64\times, 256\times$ computation for $\texttt{conv2}$-$\texttt{conv5}$, respectively). Given computational resources, brute-force computation solves shift-invariance.

\noindent {\bf Average accuracy across spatial positions}
In Figure~\ref{fig:acc_vs_pos}, we train without augmentation, and show how accuracy systematically degrades as a function of spatial shift. We observe the following:
\vspace{-3mm}
\begin{itemize}[noitemsep, leftmargin=2.5mm]
	\item On the left, the baseline heatmap shows that classification accuracy holds when testing with no shift, but quickly degrades when shifting.
	\item The proposed filtering decreases the degradation. \textit{\textit{Bin-7}} is largely consistent across all spatial positions.
	\item On the right, we plot the accuracy when making diagonal shifts to the input. As increased filtering is added, classification accuracy becomes consistent in all positions.
\end{itemize}

\noindent {\bf Classification variation distribution}
The consistency metric in the main paper looks at the hard classification, discounting classifier confidence. Similar to~\citet{azulay2019deep}, we also compute the variation in probability of correct classification (the traces shown in Figure~\ref{fig:fig1} in the main paper), given different shifts. We can capture the variation across all possible shifts: $\sqrt{ Var_{h,w}( \{{P_{\text{correct class}} (\Shift _{h,w}(X)) \}}  ) }$.

In Figure~\ref{fig:variation_change}, we show the distribution of classification variations, before and after adding in the low-pass filter. Even with a small $2\times 2$ filter, immediately variation decreases. As the filter size is increased, the output classification variation continues to decrease. This has a larger effect when training without data augmentation, but is still observable when training with data augmentation.

Training with data augmentation with the baseline network reduces variation. Anti-aliasing the networks reduces variation in both scenarios. More aggressive filtering further decreases variation.

\noindent {\bf Robustness to shift-based adversary}
In the main paper, we show that anti-aliased the networks increases the classification consistency, while maintaining accuracy. A logical consequence is increased accuracy in presence of a shift-based adversary. We empirically confirm this in Figure~\ref{fig:shift-adv} for VGG13 on CIFAR10. We compute classification accuracy as a function of maximum adversarial shift. A max shift of 2 means the adversary can choose any of the 25 positions within a $5\times 5$ window. For the classifier to ``win", it must correctly classify all of them correctly. Max shift of 0 means that there is no adversary. Conversely, a max shift of 16 means the image must be correctly classified at all $32\times32=1024$ positions.

Our primary observations are as follows:
\begin{itemize}[noitemsep,leftmargin=2.5mm]
	\item As seen in Figure~\ref{fig:shift-adv} (left), the baseline network (\textcolor{gray}{gray}) is very sensitive to the adversary.
	\item Adding larger \textit{Binomial} filters (from red to purple) increases robustness to the adversary. In fact, \textit{Bin-7} filter (purple) \textit{without} augmentation outperforms the baseline (black) \textit{with} augmentation.
	\item As seen in Figure~\ref{fig:shift-adv} (right), adding larger \textit{Binomial} filters also increases adversarial robustness, even when training with augmentation.
\end{itemize}

These results corroborate the findings in the main paper, and demonstrate a use case: increased robustness to a shift-based adversarial attack.

\subsection{Alternatives to MaxBlurPool}

In the paper, we follow signal processing first principles, to arrive at our solution of MaxBlurPool, with a fixed blurring kernel. Here, we explore possible alternatives -- swapping max and blur operations, combining max and blur in parallel through soft-gating, and learning the blur filter.

\noindent {\bf Swapping max and blur} We blur after max, immediately before subsampling, which has solid theoretical backing in sampling theory. What happens when the operations are swapped? The signal before the max operator is undoubtedly related to the signal after. Thus, blurring before max provides ``second-hand" anti-aliasing and still increases shift-invariance over the baseline. However, switching the order is worse than max and blurring in the correct, proposed order.
For example, for \textit{Bin-7}, accuracy ($93.2\rightarrow 92.6$) and consistency ($98.8\rightarrow 98.6$) both decrease.
We consistently observe this across filters. 

\noindent {\bf Softly gating between max-pool and average-pool} \citet{lee2016generalizing} investigate combining MaxPool and AvgPool in parallel, with a soft-gating mechanism, called ``Mixed'' Max-AvgPool. We instead combine them in series.
We conduct additional experiments here. On CIFAR (VGG w/ aug, see Tab~\ref{tab:results-cifar-supp}), MixedPool can offer improvements over MaxPool baseline (96.6$\rightarrow$97.2 consistency). However, by softly weighting AvgPool, some antialiasing capability is left on the table. MaxBlurPool provides higher invariance (97.6). All have similar accuracy -- 93.8, 93.7, and 93.7 for baseline MaxPool, MixedPool, and our MaxBlurPool, respectively. We use our \textit{Rect-2} variant here for clean comparison.

Importantly, our paper proposes a methodology, not a pooling layer. The same technique to modify MaxPool (reduce stride, then BlurPool) applies to the MixedPool layer, increasing its shift-invariance (97.2$\rightarrow$97.8).

\noindent {\bf Learning the blur filter} We have shown that adding anti-aliasing filtering improves shift-equivariance. What if the blur kernel were learned? We initialize the filters with our fixed weights, \textit{Tri-3} and \textit{Bin-5}, and allow them to be adjusted during training (while constraining the kernel symmetrical). The function space has more degrees of freedom and is strictly more general. However, we find that while accuracy holds, consistency decreases: relative to the fixed filters, we see $98.0\rightarrow 97.5$ for length-3 and $98.4 \rightarrow 97.3$ for length-5. While shift-invariance \textit{can} be learned, there is no explicit incentive to do so. Analogously, a fully connected network \textit{can} learn convolution, but does not do so in practice.

\section{ImageNet Classification}

We show expanded results and visualizations.

\noindent {\bf Classification and shift-invariance results} In Table~\ref{tab:results-imagenet-supp}, we show expanded results. These results are plotted in Figure~\ref{fig:consist_vs_acc_imagenet_agg} in the main paper. All pretrained models are available at \url{https://richzhang.github.io/antialiased-cnns/}.

\noindent {\bf Robustness results} In the main paper, we show aggregated results for robustness tests on the Imagenet-C/P datasets~\cite{hendrycks2019using}. In Tables~\ref{tab:results-imagenetp} and ~\ref{tab:results-imagenetc} we show expanded results, separated by each corruption and perturbation type.

Antialiasing is motivated by shift-invariance. Indeed, using the \textit{Bin-5} antialiasing filter reduces flip rate by $22.3\%$ to translations. Table~\ref{tab:results-imagenetp} indicates increased stability to other perturbation types as well. We observe higher stability to geometric perturbations -- rotation, tilting, and scaling. In addition, antialiasing also helps stability to noise. This is somewhat expected, as adding low-pass filtering helps can average away spurious noise. Surprisingly, adding blurring within the network also increases resilience to blurred images. In total, antialiasing increases stability almost across the board -- 9 of the 10 perturbations are reliably stabilized.

We also observe increased accuracy, in the face of corruptions, as shown in Table~\ref{tab:results-imagenetc}. Again, adding low-pass filtering helps smooth away spurious noise on the input, helping better maintain performance. Other high-frequency perturbations, such as pixelation and jpeg compression, are also consistency improved with antialiasing. Overall, antialiasing increases robustness to perturbations -- 13 of the 15 corruptions are reliably improved.

In total, these results indicate that adding antialiasing provides a smoother feature extractor, which is more stable and robust to out-of-distribution perturbations.

\begin{table*}[t]
\scalebox{.9} {
 \begin{tabular}{c c c c c c c c c c c c c}
 \toprule

& \multicolumn{4}{c}{\bf AlexNet} & \multicolumn{4}{c}{\bf VGG16} & \multicolumn{4}{c}{\bf VGG16bn}
\\ \cmidrule(lr){2-5} \cmidrule(lr){6-9} \cmidrule(lr){10-13}

{\bf Filter} & \multicolumn{2}{c}{\bf Accuracy} & \multicolumn{2}{c}{\bf Consistency} & \multicolumn{2}{c}{\bf Accuracy} & \multicolumn{2}{c}{\bf Consistency} & \multicolumn{2}{c}{\bf Accuracy} & \multicolumn{2}{c}{\bf Consistency} \\ \cmidrule(lr){2-3} \cmidrule(lr){4-5} \cmidrule(lr){6-7} \cmidrule(lr){8-9} \cmidrule(lr){10-11} \cmidrule(lr){12-13} 

& {\bf Abs} & {\bf $\Delta$} & {\bf Abs} & {\bf $\Delta$} & {\bf Abs} & {\bf $\Delta$} & {\bf Abs} & {\bf $\Delta$} & {\bf Abs} & {\bf $\Delta$} & {\bf Abs} & {\bf $\Delta$} \\ \midrule

{\bf Baseline}
& 56.55 & -- & 78.18 & -- 
& 71.59 & -- & 88.52 & -- 
& 73.36 & -- & 89.24 & -- 
\\
\cdashline{1-13}
{\bf Rect-2}
& {\bf 57.24} & {\bf +0.69} & 81.33 & +3.15 
& 72.15 & +0.56 & 89.24 & +0.72 
& 74.01 & +0.65 & 90.72 & +1.48 
\\
{\bf Tri-3}
& 56.90 & +0.35 & 82.15 & +3.97 
& 72.20 & +0.61 & 89.60 & +1.08 
& 73.91 & +0.55 & 91.10 & +1.86 
\\
{\bf Bin-5}
& 56.58 & +0.03 & {\bf 82.51} & {\bf +4.33} 
& {\bf 72.33} & {\bf +0.74} & {\bf 90.19} & {\bf +1.67} 
& {\bf 74.05} & {\bf +0.69} & {\bf 91.35} & {\bf +2.11} 
\\

\bottomrule

\\
\toprule

& \multicolumn{4}{c}{\bf ResNet18} & \multicolumn{4}{c}{\bf ResNet34} & \multicolumn{4}{c}{\bf ResNet50}
\\ \cmidrule(lr){2-5} \cmidrule(lr){6-9} \cmidrule(lr){10-13}

{\bf Filter} & \multicolumn{2}{c}{\bf Accuracy} & \multicolumn{2}{c}{\bf Consistency} & \multicolumn{2}{c}{\bf Accuracy} & \multicolumn{2}{c}{\bf Consistency} & \multicolumn{2}{c}{\bf Accuracy} & \multicolumn{2}{c}{\bf Consistency} \\ \cmidrule(lr){2-3} \cmidrule(lr){4-5} \cmidrule(lr){6-7} \cmidrule(lr){8-9} \cmidrule(lr){10-11} \cmidrule(lr){12-13} 

& {\bf Abs} & {\bf $\Delta$} & {\bf Abs} & {\bf $\Delta$} & {\bf Abs} & {\bf $\Delta$} & {\bf Abs} & {\bf $\Delta$} & {\bf Abs} & {\bf $\Delta$} & {\bf Abs} & {\bf $\Delta$} \\ \midrule

{\bf Baseline}
& 69.74 & -- & 85.11 & -- 
& 73.30 & -- & 87.56 & -- 
& 76.16 & -- & 89.20 & -- 
\\
\cdashline{1-13}
{\bf Rect-2}
& 71.39 & +1.65 & 86.90 & +1.79 
& {\bf 74.46} & {\bf +1.16} & 89.14 & +1.58 
& 76.81 & +0.65 & 89.96 & +0.76 
\\
{\bf Tri-3}
& {\bf 71.69} & {\bf +1.95} & 87.51 & +2.40 
& 74.33 & +1.03 & 89.32 & +1.76 
& 76.83 & +0.67 & 90.91 & +1.71 
\\
{\bf Bin-5}
& 71.38 & +1.64 & {\bf 88.25} & {\bf +3.14} 
& 74.20 & +0.90 & {\bf 89.49} & {\bf +1.93} 
& {\bf 77.04} & {\bf +0.88} & {\bf 91.31} & {\bf +2.11} 
\\

\bottomrule

\\
\toprule

& \multicolumn{4}{c}{\bf ResNet101} & \multicolumn{4}{c}{\bf DenseNet121} & \multicolumn{4}{c}{\bf MobileNetv2}
\\ \cmidrule(lr){2-5} \cmidrule(lr){6-9} \cmidrule(lr){10-13}

{\bf Filter} & \multicolumn{2}{c}{\bf Accuracy} & \multicolumn{2}{c}{\bf Consistency} & \multicolumn{2}{c}{\bf Accuracy} & \multicolumn{2}{c}{\bf Consistency} & \multicolumn{2}{c}{\bf Accuracy} & \multicolumn{2}{c}{\bf Consistency} \\ \cmidrule(lr){2-3} \cmidrule(lr){4-5} \cmidrule(lr){6-7} \cmidrule(lr){8-9} \cmidrule(lr){10-11} \cmidrule(lr){12-13} 

& {\bf Abs} & {\bf $\Delta$} & {\bf Abs} & {\bf $\Delta$} & {\bf Abs} & {\bf $\Delta$} & {\bf Abs} & {\bf $\Delta$} & {\bf Abs} & {\bf $\Delta$} & {\bf Abs} & {\bf $\Delta$} \\ \midrule

{\bf Baseline}
& 77.37 & -- & 89.81 & -- 
& 74.43 & -- & 88.81 & -- 
& 71.88 & -- & 86.50 & -- 
\\
\cdashline{1-13}
{\bf Rect-2}
& 77.82 & +0.45 & 91.04 & +1.23 
& 75.04 & +0.61 & 89.53 & +0.72 
& {\bf 72.63} & {\bf +0.75} & 87.33 & +0.83 
\\
{\bf Tri-3}
& {\bf 78.13} & {\bf +0.76} & 91.62 & +1.81 
& {\bf 75.14} & {\bf +0.71} & 89.78 & +0.97 
& 72.59 & +0.71 & 87.46 & +0.96 
\\
{\bf Bin-5}
& 77.92 & +0.55 & {\bf 91.74} & {\bf +1.93} 
& 75.03 & +0.60 & {\bf 90.39} & {\bf +1.58} 
& 72.50 & +0.62 & {\bf 87.79} & {\bf +1.29} 
\\

\bottomrule

\vspace{-5mm}
\end{tabular}
}
\vspace{-5mm}
\caption{{\bf Imagenet Classification.} We show 1000-way classification accuracy and consistency (higher is better), across 4 architectures, with anti-aliasing filtering added. We test 3 possible filters, in addition to the off-the-shelf reference models. This shows results plotted in Figure~\ref{fig:consist_vs_acc_imagenet_agg} in the main paper. {\bf Abs} is the absolute performance, and {\bf $\Delta$} is the difference to the baseline. As designed, classification consistency is improved across all methods. Interestingly, accuracy is \textit{also improved}.
\label{tab:results-imagenet-supp}
}
\end{table*}

\begin{table*}[t]
\scalebox{.7} {
 \begin{tabular}{c c c c c c c c c c c c c c c c c c}
& \multicolumn{17}{c}{\bf ResNet50 on ImageNet-C~\cite{hendrycks2019using}} \\ 
 \toprule
& \multicolumn{17}{c}{\bf Corruption Error (CE) (lower is better)} \\ \cmidrule(lr){2-18}
& \multicolumn{3}{c}{\bf Noise} & \multicolumn{4}{c}{\bf Blur} & \multicolumn{4}{c}{\bf Weather} & \multicolumn{4}{c}{\bf Digital} & \multicolumn{2}{c}{\bf Mean}
\\ \cmidrule(lr){2-4} \cmidrule(lr){5-8} \cmidrule(lr){9-12} \cmidrule(lr){13-16} \cmidrule(lr){17-18}
& {\bf Gauss} & {\bf Shot} & {\bf Impulse} & {\bf Defocus} & {\bf Glass} & {\bf Motion} & {\bf Zoom} & {\bf Snow} & {\bf Frost} & {\bf Fog} & {\bf Bright} & {\bf Contrast} & {\bf Elastic} & {\bf Pixel} & {\bf Jpeg} & {\bf Unnorm} & {\bf Norm}
\\ \midrule
{\bf Baseline} &  68.70 & 71.10 & 74.04 & 61.40 & 73.39 & 61.43 & 63.93 & 67.76 & 62.08 & 54.61 & 32.04 & 61.25 & {\bf 55.24} & 55.24 & 46.32 & 60.57 & 76.43 \\ \cdashline{1-18}
{\bf Rect-2} &  65.81 & 68.27 & 70.49 & 60.01 & 72.14 & 62.19 & 63.96 & 68.00 & 61.83 & 54.95 & 32.09 & 60.25 & 55.56 & 53.89 & 43.62 & 59.54 & 75.16 \\
{\bf Tri-3} & {\bf 63.86} & {\bf 66.07} & {\bf 69.15} & {\bf 58.36} & 71.70 & {\bf 60.74} & 61.58 & {\bf 66.78} & 60.29 & 54.40 & {\bf 31.48} & {\bf 58.09} & 55.26 & 53.89 & 43.62 & 58.35 & 73.73 \\
{\bf Bin-5} & 64.31 & 66.39 & 69.88 & 60.31 & {\bf 71.37} & 61.60 & {\bf 61.25} & 66.82 & {\bf 59.82} & {\bf 51.84} & 31.51 & 58.12 & 55.29 & {\bf 50.81} & {\bf 42.84} & {\bf 58.14} & {\bf 73.41} \\
\bottomrule

\\ \toprule
& \multicolumn{17}{c}{\bf Corruption Error, Percentage reduced from Baseline ResNet50 (higher is better)} \\ \cmidrule(lr){2-18}
& \multicolumn{3}{c}{\bf Noise} & \multicolumn{4}{c}{\bf Blur} & \multicolumn{4}{c}{\bf Weather} & \multicolumn{4}{c}{\bf Digital} & \multicolumn{2}{c}{\bf Mean}
\\ \cmidrule(lr){2-4} \cmidrule(lr){5-8} \cmidrule(lr){9-12} \cmidrule(lr){13-16} \cmidrule(lr){17-18}
& {\bf Gauss} & {\bf Shot} & {\bf Impulse} & {\bf Defocus} & {\bf Glass} & {\bf Motion} & {\bf Zoom} & {\bf Snow} & {\bf Frost} & {\bf Fog} & {\bf Bright} & {\bf Contrast} & {\bf Elastic} & {\bf Pixel} & {\bf Jpeg} & {\bf Unnorm} & {\bf Norm}
\\ \midrule
{\bf Baseline} &  0.00 & 0.00 & 0.00 & 0.00 & 0.00 & 0.00 & 0.00 & 0.00 & 0.00 & 0.00 & 0.00 & 0.00 & {\bf 0.00} & 0.00 & 0.00 & 0.00 & 0.00 \\ \cdashline{1-18}
{\bf Rect-2} &  4.21 & 3.98 & 4.79 & 2.26 & 1.70 & -1.24 & -0.05 & -0.35 & 0.40 & -0.62 & -0.16 & 1.63 & -0.58 & 2.44 & 5.83 & 1.62 & 1.32 \\
{\bf Tri-3} & {\bf 7.05} & {\bf 7.07} & {\bf 6.60} & {\bf 4.95} & 2.30 & {\bf 1.12} & 3.68 & {\bf 1.45} & 2.88 & 0.38 & {\bf 1.75} & {\bf 5.16} & -0.04 & 2.44 & 5.83 & 3.51 & 3.34 \\
{\bf Bin-5} & 6.39 & 6.62 & 5.62 & 1.78 & {\bf 2.75} & -0.28 & {\bf 4.19} & 1.39 & {\bf 3.64} & {\bf 5.07} & 1.65 & 5.11 & -0.09 & {\bf 8.02} & {\bf 7.51} & {\bf 3.96} & {\bf 3.70} \\
\bottomrule

\vspace{-5mm}
\end{tabular}
}
\vspace{-5mm}
\caption{{\bf Generalization to Corruptions.} {\bf (Top)} Corruption error rate (lower is better) of Resnet50 on the Imagenet-C. With antialiasing, the error rate decreases, often times significantly, on most corruptions. {\bf (Bottom)} The percentage reduction relative to the baseline ResNet50 (higher is better). The right two columns show mean across corruptions. ``Unnorm'' is the raw average. ``Norm'' is normalized to errors made from AlexNet, as proposed in ~\cite{hendrycks2019using}.
\vspace{-4mm}
\label{tab:results-imagenetc}
}
\end{table*}

\begin{table*}[t]
\scalebox{.8} {
 \begin{tabular}{c c c c c c c c c c c c c}
 & \multicolumn{12}{c}{\bf ResNet50 on ImageNet-P~\cite{hendrycks2019using}} \\ 
 \toprule
& \multicolumn{12}{c}{\bf Flip Rate (FR) (lower is better)} \\ \cmidrule(lr){2-13}
& \multicolumn{2}{c}{\bf Noise} & \multicolumn{2}{c}{\bf Blur} & \multicolumn{2}{c}{\bf Weather} & \multicolumn{4}{c}{\bf Geometric} & \multicolumn{2}{c}{\bf Mean} \\ \cmidrule(lr){2-3} \cmidrule(lr){4-5} \cmidrule(lr){6-7} \cmidrule(lr){8-11} \cmidrule(lr){12-13}
& {\bf Gauss} & {\bf Shot} & {\bf Motion} & {\bf Zoom} & {\bf Snow} & {\bf Bright} & {\bf Translate} & {\bf Rotate} & {\bf Tilt} & {\bf Scale} & {\bf Unnorm} & {\bf Norm} \\ \midrule
{\bf Baseline} & 14.04 & 17.38 & 6.00 & 4.29 & 7.54 & 3.03 & 4.86 & 6.79 & 4.01 & 11.32 & 7.92 & 57.99 \\ \cdashline{1-13}
{\bf Rect-2} & 14.08 & 17.16 & 5.98 & 4.21 & 7.34 & 3.20 & 4.42 & 6.43 & 3.80 & 10.61 & 7.72 & 56.70 \\
{\bf Tri-3} & 12.59 & 15.57 & {\bf 5.39} & 3.79 & 6.98 & {\bf 3.01} & 3.95 & 5.80 & 3.53 & 9.90 & 7.05 & 51.91 \\
{\bf Bin-5} & {\bf 12.39} & {\bf 15.22} & 5.44 & {\bf 3.72} & {\bf 6.76} & 3.15 & {\bf 3.78} & {\bf 5.67} & {\bf 3.44} & {\bf 9.45} & {\bf 6.90} & {\bf 51.18} \\

\bottomrule

\\ \toprule
& \multicolumn{12}{c}{\bf Flip Rate (FR) [Percentage reduced from Baseline] (higher is better)} \\ \cmidrule(lr){2-13}
& \multicolumn{2}{c}{\bf Noise} & \multicolumn{2}{c}{\bf Blur} & \multicolumn{2}{c}{\bf Weather} & \multicolumn{4}{c}{\bf Geometric} & \multicolumn{2}{c}{\bf Mean} \\ \cmidrule(lr){2-3} \cmidrule(lr){4-5} \cmidrule(lr){6-7} \cmidrule(lr){8-11} \cmidrule(lr){12-13}
& {\bf Gauss} & {\bf Shot} & {\bf Motion} & {\bf Zoom} & {\bf Snow} & {\bf Bright} & {\bf Translate} & {\bf Rotate} & {\bf Tilt} & {\bf Scale} & {\bf Unnorm} & {\bf Norm} \\ \midrule
{\bf Baseline} & 0.00 & 0.00 & 0.00 & 0.00 & 0.00 & 0.00 & 0.00 & 0.00 & 0.00 & 0.00 & 0.00 & 0.00 \\ \cdashline{1-13}
{\bf Rect-2} & -0.25 & 1.27 & 0.30 & 1.73 & 2.65 & -5.75 & 9.21 & 5.34 & 5.16 & 6.20 & 2.55 & 2.22 \\
{\bf Tri-3} & 10.35 & 10.41 & {\bf 10.09} & 11.58 & 7.42 & {\bf 0.53} & 18.89 & 14.55 & 12.02 & 12.50 & 11.03 & 10.48 \\
{\bf Bin-5} & {\bf 11.81} & {\bf 12.42} & 9.27 & {\bf 13.28} & {\bf 10.28} & -4.10 & {\bf 22.27} & {\bf 16.59} & {\bf 14.11} & {\bf 16.50} & {\bf 12.91} & {\bf 11.75} \\

\bottomrule

\\ \toprule
& \multicolumn{12}{c}{\bf Top-5 Distance (T5D) (lower is better)} \\ \cmidrule(lr){2-13}
& \multicolumn{2}{c}{\bf Noise} & \multicolumn{2}{c}{\bf Blur} & \multicolumn{2}{c}{\bf Weather} & \multicolumn{4}{c}{\bf Geometric} & \multicolumn{2}{c}{\bf Mean} \\ \cmidrule(lr){2-3} \cmidrule(lr){4-5} \cmidrule(lr){6-7} \cmidrule(lr){8-11} \cmidrule(lr){12-13}
& {\bf Gauss} & {\bf Shot} & {\bf Motion} & {\bf Zoom} & {\bf Snow} & {\bf Bright} & {\bf Translate} & {\bf Rotate} & {\bf Tilt} & {\bf Scale} & {\bf Unnorm} & {\bf Norm} \\ \midrule
{\bf Baseline} & 3.92 & 4.55 & 1.63 & 1.20 & 1.95 & {\bf 1.00} & 1.68 & 2.15 & 1.40 & 3.01 & 2.25 & 78.36 \\ \cdashline{1-13}
{\bf Rect-2} & 3.94 & 4.54 & 1.63 & 1.19 & 1.91 & 1.06 & 1.56 & 2.07 & 1.34 & 2.89 & 2.21 & 77.40 \\
{\bf Tri-3} & 3.67 & 4.28 & {\bf 1.50} & 1.10 & 1.85 & 1.00 & 1.43 & 1.92 & 1.25 & 2.72 & 2.07 & 72.36 \\
{\bf Bin-5} & {\bf 3.65} & {\bf 4.22} & 1.53 & {\bf 1.09} & {\bf 1.78} & 1.04 & {\bf 1.39} & {\bf 1.89} & {\bf 1.25} & {\bf 2.66} & {\bf 2.05} & {\bf 71.86} \\

\bottomrule

\\ \toprule
& \multicolumn{12}{c}{\bf Top-5 Distance (T5D) [Percentage reduced from Baseline] (higher is better)} \\ \cmidrule(lr){2-13}
& \multicolumn{2}{c}{\bf Noise} & \multicolumn{2}{c}{\bf Blur} & \multicolumn{2}{c}{\bf Weather} & \multicolumn{4}{c}{\bf Geometric} & \multicolumn{2}{c}{\bf Mean} \\ \cmidrule(lr){2-3} \cmidrule(lr){4-5} \cmidrule(lr){6-7} \cmidrule(lr){8-11} \cmidrule(lr){12-13}
& {\bf Gauss} & {\bf Shot} & {\bf Motion} & {\bf Zoom} & {\bf Snow} & {\bf Bright} & {\bf Translate} & {\bf Rotate} & {\bf Tilt} & {\bf Scale} & {\bf Unnorm} & {\bf Norm} \\ \midrule
{\bf Baseline} & 0.00 & 0.00 & 0.00 & 0.00 & 0.00 & {\bf 0.00} & 0.00 & 0.00 & 0.00 & 0.00 & 0.00 & 0.00 \\ \cdashline{1-13}
{\bf Rect-2} & -0.41 & 0.09 & -0.12 & 0.39 & 1.71 & -5.83 & 7.19 & 3.74 & 3.90 & 3.93 & 1.51 & 1.22 \\
{\bf Tri-3} & 6.53 & 5.82 & {\bf 7.95} & 8.10 & 5.21 & -0.65 & 15.11 & 10.82 & 10.26 & 9.80 & 7.86 & 7.65 \\
{\bf Bin-5} & {\bf 7.03} & {\bf 7.26} & 6.24 & {\bf 9.15} & {\bf 8.45} & -4.13 & {\bf 17.73} & {\bf 12.15} & {\bf 10.62} & {\bf 11.80} & {\bf 8.91} & {\bf 8.30} \\

\bottomrule

\vspace{-5mm}
\end{tabular}
}
\vspace{-5mm}
\caption{{\bf Stability to Perturbations.} Flip Rate (FR) and Top-5 Distance (T5D) of ResNet50 on ImageNet-P. Though our antialiasing is motivated by shift-invariance (``translate''), it adds additional stability across many other perturbation types.
\vspace{-4mm}
\label{tab:results-imagenetp}
}
\end{table*}

\section{Qualitative examples for Labels$\rightarrow$Facades}

In the main paper, we discussed the tension between needing to generate high-frequency content and low-pass filtering for shift-invariance. Here, we show an example of applying increasingly aggressive filters. In general, generation quality is maintained with the \textit{Rect-2} and \textit{Tri-3} filters, and then degrades with additional filtering.

\begin{figure*}
\centering
\includegraphics[width=1.\linewidth,trim={0cm 0.2cm 0cm 0cm}, clip]{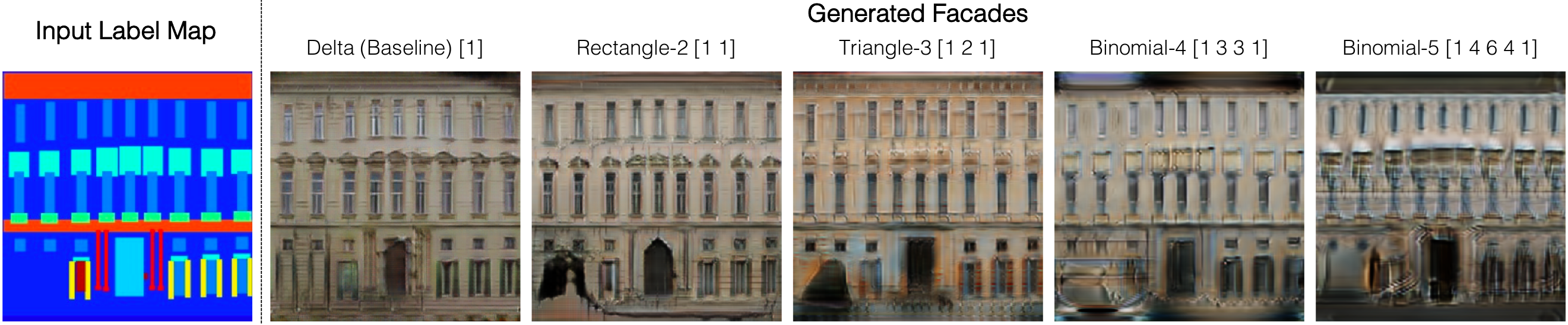}
\vspace{-10mm}
\caption{
{\bf Example generations.} We show generations with U-Nets trained with 5 different filters. In general, generation quality is well-maintained to \textit{\textbf{Tri-3}} filter, but decreases noticeably with \textit{\textbf{ Bin-4}} and \textit{\textbf{Bin-5}} filters due to oversmoothing.
\label{fig:gen_example}
\vspace{-2mm}
}
\end{figure*}



\end{document}


\twocolumn[
\icmltitle{Making Convolutional Networks Shift-Invariant Again \\ (Supplementary Material)}




\begin{icmlauthorlist}
\icmlauthor{Richard Zhang}{adobe}
\end{icmlauthorlist}

\icmlaffiliation{adobe}{Adobe Research, San Francisco, CA, US}

\icmlcorrespondingauthor{Richard Zhang}{rizhang@adobe.com}

\icmlkeywords{Convolutional networks, shift-invariance, anti-aliasing}

\vskip 0.3in
]




We show additional qualitative examples for image-to-image translation task in the attached website and video. We show additional quantitative analysis for VGG classification in this document.

\section{Qualitative examples for Labels$\rightarrow$Facades}

Attached is a website \texttt{website/index.html}, which contains additional examples of generated facades. This is an extension of Figure 7 in the main paper. The examples are randomly selected from the test set. We find that quality degrades for \textit{\textbf{Bin-4}} and \textit{\textbf{Bin-5}}.

Also attached are videos showing generated facades, given the same input map, but shifted (see \texttt{videos} directory). This is an extension of Figure 6 in the main paper. The generated images are from \textit{\textbf{Delta-1}} (baseline), \textit{\textbf{Rect-2}}, \textit{\textbf{Tri-3}}, \textit{\textbf{Bin-4}}, \textit{\textbf{Bin-5}} filters. Note that with the baseline method, features such as windows shift with the convolutional grid. With filtering added, features remain more static.

\section{Quantitative Classification Analysis}

\subsection{How do the learned convolutional filters change?}

Our proposed change smooths the internal feature maps for purposes of downsampling. How does training with this layer affect the \textit{learned} convolutional layers. We measure spatial smoothness using the normalized Total Variation (TV) metric proposed in~\cite{ruderman2018pooling}. 
A higher value indicates a filter with more high-frequency components. A lower value indicates a smoother filter. As shown in Fig.~\ref{fig:filt_tv}, using the MaxBlurPool (red-purple) actually induces smoother \textit{learned} filters throughout the network, relative to the MaxPool baseline (black). Adding in more aggressive blur kernels further decreases the TV (increasing smoothness). This indicates that our method actually induces a smoother feature extractor overall.

\begin{figure}[h]
  \begin{center}
	\includegraphics[width=1.\columnwidth,trim={0.2cm 0 0.3cm 0.1cm},clip]{figures/filter_tv_2.pdf}
  \end{center}
  \vspace{-5mm}
  \caption{{\bf Total Variation (TV) by layer.} We compute average smoothness of learned conv filters per layer (lower is smoother). Baseline MaxPool is in black, and adding additional blurring is shown in colors. Note that the \textit{learned} convolutional layers become smoother, indicating that a smoother feature extractor is induced.
  \label{fig:filt_tv}
}
\end{figure}

\subsection{Average accuracy across spatial positions}

In Figure~\ref{fig:acc_vs_pos}, we show how accuracy systematically degrades as a function of spatial shift, when training without augmentation. We observe the following:

\begin{itemize}[noitemsep, leftmargin=4.5mm]
	\item On the left, the baseline heatmap shows that classification accuracy when testing with no shift, but quickly degrades when shifting.
	\item The proposed filtering decreases the degradation. \textit{\textit{Bin-7}} is largely consistent across all spatial positions.
	\item On the right, we plot the accuracy when making diagonal shifts. As increased filtering is added, classification accuracy becomes consistent in all positions.
\end{itemize}

\begin{figure*}[h]
\centering
\includegraphics[height=4.3cm, trim={0 0 0 0}, clip]{figures/rebutt_acc_vs_pos_heat.pdf}
\includegraphics[height=4.3cm, trim={10.5cm 0 0 0}, clip]{figures/rebutt_acc_vs_pos_heat_bar.pdf}
\includegraphics[height=4.5cm, trim={0.1cm 0 0 0}, clip]{figures/rebutt_acc_vs_pos2.pdf}

\caption{\textbf{Average accuracy as a function of shift.} {\bf (Left)} We show classification accuracy across the test set as a function of shift, given different filters. {\bf (Right)} We plot accuracy vs diagonal shift in the input image, across different filters. Note that accuracy degrades quickly with the baseline, but as increased filtering is added, classifications become consistent across spatial positions.
\label{fig:acc_vs_pos}
\vspace{-2mm}
}
\end{figure*}

\subsection{Classification variation distribution}

\subfile{tables/var_dist.tex}

The consistency metric in the main paper looks at the hard classification, discounting classifier confidence. Similar to ~\cite{azulay2019why}, we also compute the variation in probability of correct classification (the traces shown in Fig. 1 in the main paper), given different shifts. We can capture the variation across all possible shifts: $\sqrt{ Var_{h,w}( \{{P_{\text{correct class}} (\Shift _{h,w}(X)) \}} \} ) }$.

In Fig.~\ref{fig:variation_change}, we show the distribution of classification variations, before and after adding in the low-pass filter. Even a small $2\times 2$ filter, immediately variation decreases. As the filter size is increased, the output classification variation continues to decrease. This has a larger effect when training without data augmentation, but is still observable when training with data augmentation.

Fig.~\ref{fig:variation_change} investigates the distribution of classification variations. Training with data augmentation with the baseline network reduces variation (black lines on both plots). Our method reduces variation in both scenarios. More aggressive filtering further decreases variation.

\subsection{Robustness to shift-based adversary}

In the main paper, we show that using the proposed MaxBlurDown method increases the classification consistency, while maintaining accuracy.
A logical consequence is increased accuracy in presence of a shift-based adversary. We empirically confirm this in Fig.~\ref{fig:shift-adv} for VGG13 on CIFAR10. We compute classification accuracy as a function of maximum adversarial shift. A max shift of 2 means the adversary can choose any of the 25 positions within a $5\times 5$ window. For the classifier to ``win", it must correctly classify all of them correctly. Max shift of 0 means that there is no adversary. Conversely, a max shift of 16 means the image must be correctly classified at all $32\times32=1024$ positions.

Our primary observations are as follows:

\begin{itemize}[noitemsep,leftmargin=4.5mm]
	\item As seen in Fig.~\ref{fig:shift-adv} (left), the baseline network (\textcolor{gray}{gray}) is very sensitive to the adversary.
	\item Adding larger \textbf{\textit{Binomial}} filters (from red to purple) increases robustness to the adversary. In fact, \textbf{\textit{Bin-7}} filter (purple) \textit{without} augmentation outperforms the baseline (black) \textit{with} augmentation.
	\item As seen in Fig.~\ref{fig:shift-adv} (right), adding larger \textit{\textbf{Binomial}} filters  also increases adversarial robustness, even when training with augmentation.
\end{itemize}

These results corroborate the findings in the main paper, and demonstrate a use case: increased robustness to shift-based adversarial attack.

\subfile{tables/rebutt_adversarial.tex}
















\bibliography{iclr2019_conference}
\bibliographystyle{icml2019}

\appendix
